%% file: arxiv-P5_RSD_Poker.tex
\pdfoutput=1
\documentclass{article}
\usepackage[T1]{fontenc}
\usepackage{iclr2027_conference}
\usepackage[T1]{fontenc}
\renewcommand{\sfdefault}{phv}
\input{math_commands.tex}

\usepackage{hyperref}
\hypersetup{hidelinks,hypertexnames=false,pdfauthor={Miaobo Hu, Shuhao Hu, Xiaobo Guo, Xin Wang, Bokun Wang, Peng Zhang, Daren Zha},pdftitle={RSD-Poker: Structure-Adaptive and Shift-Robust Risk--Utility Certification}}
\usepackage{url}
\usepackage{graphicx}
\usepackage{booktabs}
\usepackage{float}
\usepackage{tabularx}
\usepackage{longtable}
\usepackage{microtype}
\usepackage{placeins}
\usepackage{xspace}
\newcommand{\best}[1]{\textbf{\boldmath #1}}
\newcommand{\second}[1]{\underline{#1}}

\title{RSD-Poker: Structure-Adaptive\\
and Shift-Robust Risk--Utility\\
Certification for Residual Policies\\
in Imperfect-Information Games}
\author{Miaobo Hu$^{1,2,*}$, Shuhao Hu$^{2}$, Xiaobo Guo$^{2}$, Xin Wang$^{2}$,\\
Bokun Wang$^{2}$, Peng Zhang$^{2}$, Daren Zha$^{2}$\\[4pt]
{\normalfont $^{1}$School of Artificial Intelligence, University of Chinese Academy of Sciences, Beijing, China}\\
{\normalfont $^{2}$Institute of Information Engineering, Chinese Academy of Sciences, Beijing, China}\\
{\normalfont $^{*}$Corresponding author: \texttt{humiaobo@iie.ac.cn}}}

\iclrfinalcopy
\begin{document}
\maketitle
\lhead{Preprint}

\begin{abstract}
Residual policy adaptation provides a lightweight way to modify a strong
reference policy, but a shared scale and a fixed subgroup partition can hide
heterogeneous degradation and become fragile when the deployment mixture of
information states changes. We introduce RSD-Poker, a structure-adaptive and
shift-robust certification framework that freezes a bank of residual families
and scales, learns a policy-visible partition on an independent structure
split, and freezes that partition before calibration labels are joined. Each
candidate--group pair receives a weighted simultaneous upper certificate for
anchor-relative risk and a lower certificate for weak-response utility. A
robust group-to-candidate map is then selected over a predeclared uncertainty
set of deployment group proportions.

Under independent calibration units drawn from each frozen group's law, a candidate
bank and partition fixed before calibration, and invariant within-group
conditionals, the selected map satisfies its declared mixture-robust risk
budget and utility certificate with probability at least
$1-\zeta_{\mathrm{risk}}-\zeta_{\mathrm{util}}$. The information contract
supports both a teacher-backed transform and a teacher-free observation-only
student. The retained deterministic 24-state audit remains an exact replay
diagnostic: empirical-zero selects $\alpha=0.08$, raising the weak-response
proxy from $4.2082$ to $4.2889$ with $0/12$ held-out threshold crossings.
On stratified held-out states, the learned-partition dual selector raises weak
utility from $4.4074$ under global dual certification to $4.4936$ and lowers
held-out violation from $0.0215$ to $0.0078$; its mixture-robust variant
reaches violation $0.0059$. Across five observation-only
checkpoints, risk-calibrated residuals attain weak utility
$4.3659\pm0.0177$ and violation rate $0.0178\pm0.0057$.
\end{abstract}

\section{Introduction}
Residual policies correct a fixed reference policy with a learned or
hand-designed adjustment \citep{silver2018residual,johannink2019residual}, and
they are often tuned and scored on the same oracle states. That
practice can make a small utility gain appear robust without showing whether
the scale survives a frozen test split. RSD-Poker asks a reproducible question:
can a residual scale be selected from calibration states, frozen, and then
audited on held-out states with a trace that records the information used to
produce each decision?

The protocol separates the observed state, the frozen teacher interface, and
the strong/weak post-trace oracle. We formulate scale selection as utility
maximization subject to a finite-sample upper bound on state-level violation
risk. The earlier zero-count rule remains an empirical baseline. Public
per-state rows bind the result to the policy distribution, source fixture,
and replay configuration. The teacher-backed transform and the observed-only
student interface have distinct information contracts: the former consumes
cached teacher distributions, while the latter receives observations and
legal actions.

State heterogeneity creates a second difficulty. A single residual scale may
be unnecessarily conservative on some information states while exceeding the
declared degradation tolerance on others. Position, pot pressure, action
history, and private information available to the acting player induce
different residual sensitivities, so aggregate risk can hide concentrated
failures in a strategically distinct subset.

A third difficulty concerns the objective used after risk certification.
Simultaneous risk bounds make post-certification selection valid for the risk
event, but they do not make the maximum calibration utility an unbiased
population estimate. We therefore certify utility with a simultaneous lower
bound and expose the group-wise candidate map in the trace.

\paragraph{From fixed groups to structure-adaptive certification.}
Coarse groups pool states with different residual sensitivities, whereas
overly fine groups reduce support and widen finite-sample certificates. We
therefore learn a bounded-depth policy-visible partition on an independent
structure split. The objective combines within-group residual heterogeneity,
minimum-support penalties, and partition complexity; the selected rule is
frozen before calibration evidence is used.

\paragraph{Robustness to deployment-mixture shift.}
Even valid group-wise bounds can be misleading when deployment reweights the
groups. We model the group-proportion change by a predeclared $\ell_1$
uncertainty set and select the map by its worst-case certified risk and
utility over that set. This covers composition shift while keeping
within-group conditional shift and adaptive-opponent evaluation as separate
endpoints.

This distinction matters in imperfect-information games. Matching the teacher's
top action, preserving its action distribution, and performing well against
an opponent are different objectives. Poker systems address strategic
uncertainty through regret minimization, search, and re-solving
\citep{zinkevich2007regret,moravcik2017deepstack,brown2018libratus}.
A residual audit instead asks what changes when a fixed distribution is tilted
by an observed-state correction. A high weak-pool score can coexist with many
state-level threshold crossings, and a zero crossing count can coexist with
a small negative mean strong delta. Reporting the distribution, denominator,
and threshold together makes these outcomes interpretable.

The primary statistical protocol uses stratified random partitions, repeated
calibration sizes, declared shifts, and five independently trained
observation-only checkpoints; a deterministic 24-state fixture supports exact
replay, and full state rows and additional student, opponent, and HUNL
evaluations appear in the appendix.

\paragraph{Contributions.}
Our contributions are fivefold. First, we define a frozen residual candidate
bank containing multiple residual directions and scales, so selection can
search a richer policy class without a monotonicity assumption. Second, we
introduce certificate-aware policy-visible partitioning: structure data
expose heterogeneous residual sensitivity while minimum support and
complexity penalties protect certificate efficiency. Third, we add weighted
simultaneous risk upper bounds and utility lower bounds whose allocations are
fixed before calibration. Fourth, we introduce mixture-robust map selection
over a declared uncertainty set of deployment group proportions. Fifth, we
make candidate construction, partition learning, calibration, map selection,
and held-out evaluation auditable under explicit trace-freeze boundaries,
with learned observation-only residuals and opponent or solver endpoints
evaluated as separate objects. Empirically, the learned-partition selector
attains the highest weak EV ($4.4936$) and utility certificate ($0.7257$) in
the stratified comparison with held-out violation $0.0078$, and the exact
fixture replay selects $\alpha=0.08$ with $0/12$ held-out threshold crossings.

\section{Related Work}
\paragraph{Game solving and policy approximation.}
Counterfactual regret minimization and its sampled and neural variants
optimize strategies in extensive-form games
\citep{zinkevich2007regret,lanctot2009mccfr,brown2019deepcfr}.
CFR+ and large-scale poker systems combine regret updates, abstraction,
search, or re-solving
\citep{tammelin2015solving,bowling2015limit,moravcik2017deepstack,brown2018libratus,
brown2020pluribus,brown2017safe,brown2020rebel}.
Neural fictitious self-play offers another route to learned mixed strategies
\citep{heinrich2016nfsp}. RSD-Poker studies a fixed teacher-backed distribution
under a scalar residual intervention; the game, opponent response model,
and scale-selection partition determine the measured result.

\paragraph{Imitation, offline learning, and constrained updates.}
Policy distillation transfers a teacher distribution to a student
\citep{rusu2015policy,czarnecki2019distilling}. Offline RL highlights the
relationship between dataset support and value estimation
\citep{kumar2020cql,kostrikov2022iql,fu2021d4rl}, while trust-region methods
control policy updates through a local optimization constraint
\citep{schulman2015trpo,schulman2017ppo}. Our evaluated operation is an explicit
exponential tilt of a cached anchor. Its scale is chosen by a finite
calibration count, and its discrepancy from the teacher is measured
separately. This separates the mechanics of shrinkage from the question of
learning a useful residual direction.

\paragraph{Finite-sample certification and adaptive structure.}
Risk-controlling prediction and learn-then-test methods calibrate a finite
family of candidate rules so that a declared loss is controlled with
finite-sample confidence \citep{angelopoulos2025learn,bates2021risk,angelopoulos2024crc}.
When calibration and evaluation distributions differ, covariate- and
label-shift corrections reweight the calibration evidence
\citep{tibshirani2019covariate,podkopaev2021labelshift}; group
distributionally robust optimization instead trains against worst-case group
mixtures \citep{sagawa2020groupdro,duchi2021dro}.
RSD-Poker uses this established calibration principle for a different
decision object: a residual-policy family with a learned policy-visible
partition in an imperfect-information game. The structure split, calibration
split, and held-out score occupy separate partitions, so the partition can
expose heterogeneous residual sensitivity without consuming the calibration
evidence. Because maximizing calibration utility can introduce tuning bias as
the candidate family grows \citep{zeng2025tuning}, our selector reports a
weighted simultaneous utility lower certificate and optimizes it over a
predeclared uncertainty set of deployment group proportions. Fixed groups and
matched mixtures are special cases of this structure-adaptive formulation,
and the robust step acts on certificates of a frozen bank rather than on
training losses.

\paragraph{Safe and constrained policy optimization.}
Constrained policy optimization maximizes reward subject to expected-cost
constraints \citep{achiam2017cpo}, and high-confidence or baseline-bootstrapped
policy improvement accepts a new policy only when it improves on a behavior
baseline with high probability \citep{thomas2015hcpi,laroche2019spibb}.
RSD-Poker uses a narrower contract:
the residual family is frozen before strong-response labels are revealed,
and calibration determines which generated intervention satisfies a declared
risk budget. This keeps the strong-response oracle out of residual training
and makes the selected scale replayable from a sealed candidate trace.

\paragraph{Poker evaluation and language-model interfaces.}
PokerBench separates action-label evaluation from game-play evaluation
\citep{zhuang2025pokerbench}; OpenSpiel provides reusable game and evaluation
interfaces \citep{lanctot2019openspiel}. AIVAT develops variance reduction for
agent evaluation under imperfect information \citep{burch2018aivat}.
We similarly distinguish historical action agreement, fixed-response EV,
and an opponent-dependent evaluation. Our percentile intervals resample the
finite fixture rows. The student interface uses distributional targets and
can be instantiated with low-rank adaptation and AdamW
\citep{hu2022lora,loshchilov2019adamw}. The appendix gives additional links between
cost-aware querying, paired simulation, and the reporting contract.

\section{Method}
\label{sec:method}
\subsection{Task definition and four-way information partition}
An observation $s$ contains the public board and betting history, the acting
player's private rank, position, pot, effective stack, and legal action set
$\mathcal A(s)$. The evaluated fixture has 24 states, four card ranks,
one public board card, and actions fold, call, and raise. The output is a
probability vector on legal actions and a deterministic top-action trace.
The evaluation unit is one state--policy pair.

We use four disjoint data roles,
\[
\mathcal D=\mathcal D_{\mathrm{train}}\cup
\mathcal D_{\mathrm{struct}}\cup\mathcal D_{\mathrm{cal}}\cup
\mathcal D_{\mathrm{test}}.
\]
The training split constructs learned residual directions when applicable.
The structure split learns the policy-visible partition and its fixed
confidence-allocation schedule. Calibration is scored only after the
candidate bank and partition are frozen, and the test split is joined only
after the final group-to-candidate map is frozen. State or independent-block
identities are disjoint across these roles.

\begin{figure}[t]
\centering
\includegraphics[width=\linewidth]{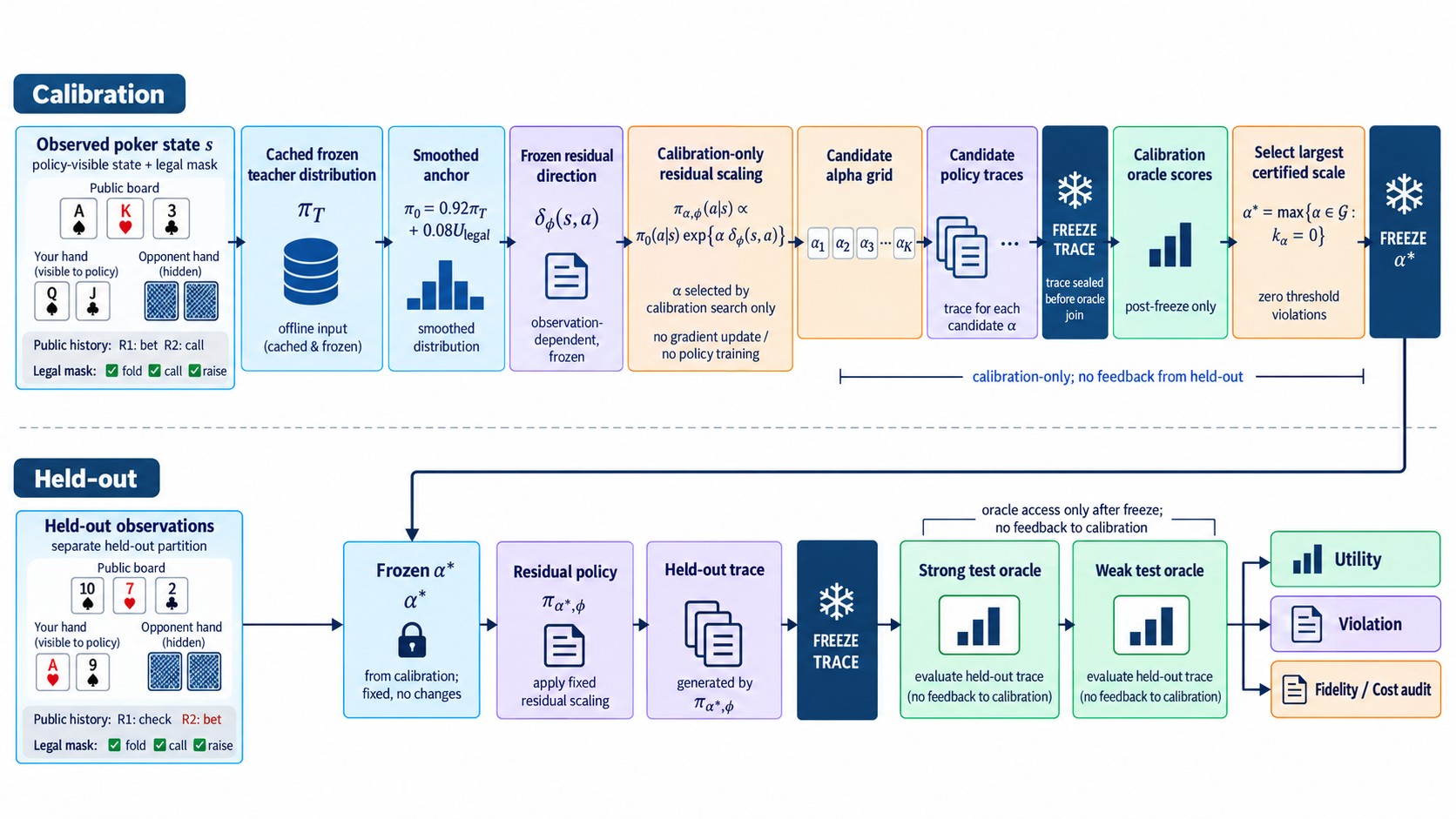}
\caption{RSD-Poker information flow for a single residual family. Top:
calibration builds the smoothed anchor from the cached teacher, forms
candidate traces over the scale grid, seals them before calibration oracle
scores are joined, and freezes the selected scale. Bottom: the frozen policy
is replayed on held-out observations, and strong, weak, and fidelity audits are
joined only after the held-out trace is frozen. Snowflakes mark freeze
boundaries; no held-out quantity feeds back into calibration.}
\label{fig:method}
\end{figure}

Four objects are kept separate. The observed record describes the policy's
available context. The teacher record supplies the cached distribution
$\pi_T(\cdot\mid s)$ used by the offline transform. The scoring record supplies
weak- and strong-response action values, joined after a trace has been sealed.
The structure record may expose response values for partition learning, but
the deployed group function receives only policy-visible features. Thus, the
teacher-backed transform has access to a teacher distribution, but
the policy input contains no strong/weak scoring vector or opponent private
card. The observed-only student contract in Appendix~\ref{app:implementation}
restricts inference to observations and legal masks. This distinction fixes
which information is available at each stage.
Figure~\ref{fig:method} shows this information flow for a single residual
family; the structure-adaptive selector replaces the scalar selection step
with the frozen group-to-candidate map of Sections~\ref{sec:groups}--\ref{sec:robust}.

\subsection{Frozen residual candidate bank}
The distributional anchor smooths the frozen teacher with a uniform legal
distribution $U_{\mathcal A(s)}$:
\begin{equation}
 \pi_0(\cdot\mid s)=0.92\pi_T(\cdot\mid s)+0.08U_{\mathcal A(s)},\qquad
 \pi_{\alpha,\phi}(a\mid s)=
 \frac{\pi_0(a\mid s)\exp\{\alpha\delta_\phi(s,a)\}}
 {\sum_{b\in\mathcal A(s)}\pi_0(b\mid s)\exp\{\alpha\delta_\phi(s,b)\}}.
 \label{eq:tilt}
\end{equation}
Here $\delta_\phi$ is a frozen residual direction. It can be a fixed
observation-dependent rule or a learned residual trained on a separate
partition. For the deterministic fixture, let
$p(s)=\min(1,\operatorname{pot}(s)/10)$ and, in legal-action order (fold,
call, raise), use
\begin{equation}
 \delta_{\mathrm{fixed}}(s)=(-0.20,\;0.05,\;0.55+p(s)).
 \label{eq:residual}
\end{equation}
Scale zero recovers the anchor and scale one is the unconstrained stress
control. Illegal actions receive zero mass and ties use the declared
legal-action order. The residual family, parameters, ordered grid, action
serializer, and candidate traces are frozen before calibration scores are
joined. Appendix~\ref{app:method} gives the complete operational definition.

More generally, let $\{\delta_{\phi_j}\}_{j=1}^{J}$ be a finite collection of
residual directions constructed without calibration or held-out strong-response
labels. The candidate bank is
\[
\mathcal B=\{b=(j,\alpha):j\in\{1,\ldots,J\},\ \alpha\in\mathcal G\},
\qquad K=|\mathcal B|,
\]
with $\pi_b$ defined by Equation~\ref{eq:tilt}. Its manifest records the
residual-family identifier, checkpoint digest where applicable, scale,
legal-action serializer, normalization rule, and complete candidate trace.
The manifest is frozen before any calibration oracle record is joined. The
fixed-fixture experiment is the special case $J=1$.

\paragraph{Bank construction chronology.}
Before the structure split is scored, the residual-family identifiers,
checkpoint digests, ordered scale grid, legal-action serializer, anchor
normalization, and candidate traces are frozen. Structure learning can use
only the separate structure records and cannot change residual parameters or
the predeclared partition class. The complete candidate bank used for
calibration is sealed before $\mathcal D_{\mathrm{cal}}$ is joined.

\subsection{Policy-visible groups and adaptive residual maps}
\label{sec:groups}
Let $\psi(s)$ contain only policy-visible attributes such as position,
pot-pressure bins, history depth, effective-stack bins, and public-board
features. We consider a predeclared class $\mathcal P$ of bounded-depth
partitions whose leaves are functions of $\psi(s)$; response values,
opponent-private variables, held-out identities, and solver outputs are
excluded from the deployed rule. For candidate $b$, let
$d_b(s)=E_s(\pi_b,s)-E_s(\pi_0,s)$ on the structure split and define
\[
\widehat\Gamma(g)=\frac{1}{K}
\sum_{b\in\mathcal B}\sum_{h\in\mathcal H(g)}
\frac{n_h^{\mathrm{struct}}}{|\mathcal D_{\mathrm{struct}}|}
\operatorname{Var}\!\left[d_b(s)\mid g(s)=h\right].
\]
The certificate-aware structure objective is
\[
\widehat{\mathcal J}_{\mathrm{part}}(g)=
\widehat\Gamma(g)+\lambda_{\mathrm{cert}}
\sum_{h\in\mathcal H(g)}\frac{1}{\sqrt{\max(n_h^{\mathrm{struct}},1)}}
+\lambda_{\mathrm{comp}}|\mathcal H(g)|,
\]
subject to $n_h^{\mathrm{struct}}\ge n_{\min}$. We select
$g^\star=\arg\min_{g\in\mathcal P}\widehat{\mathcal J}_{\mathrm{part}}(g)$,
breaking ties by fewer leaves and then deterministic rule serialization, and
freeze it before calibration. A singleton $\mathcal P$ recovers the fixed
group baseline.

Let $g^\star:\mathcal S\rightarrow\mathcal H$ be the resulting mapping from
an observed information state to one of $H=|\mathcal H|$ disjoint groups.
For candidate $b$ and state $s$, define
\[
d_b(s)=E_s(\pi_b,s)-E_s(\pi_0,s),\qquad
v_b(s)=\mathbf 1\{d_b(s)<-\tau\}.
\]
The deployed map may select one candidate per observable group,
\[
\pi^\star(\cdot\mid s)=\pi_{b^\star_{g^\star(s)}}(\cdot\mid s),
\]
with the anchor as an explicit fallback when no candidate is certified.
The partition, ordering, bank, scales, budgets, confidence weights, and
fallback rule are fixed before calibration scores are revealed.

The utility evaluator defines the optimization objective; the risk evaluator
defines unacceptable degradation relative to the anchor. In the poker
instantiation these are the weak- and strong-response evaluators.
For pool $g\in\{w,s\}$, the post-trace score is
$E_g(\pi,s)=\sum_a\pi(a\mid s)Q_g(s,a)$. The expected-distribution score
and the top action measure different aspects of a policy. A residual can
change $E_g$ while leaving the top action unchanged. Define the
anchor-relative strong degradation and its threshold indicator as
\begin{equation}
 d_\alpha(s)=E_s(\pi_{\alpha,\phi},s)-E_s(\pi_0,s),\qquad
 v_\alpha(s)=\mathbf 1\{d_\alpha(s)<-\tau\},
 \label{eq:violation}
\end{equation}
where $\tau>0$ is fixed before calibration. The inequality is strict and
all aggregates retain their declared state denominator.

\subsection{Dual risk--utility certification}
For group $h$, let $\mathcal C_h=\{s_i\in\mathcal C:g(s_i)=h\}$ and
$n_h=|\mathcal C_h|$. The violation count for candidate $b$ is
$k_{b,h}=\sum_{s_i\in\mathcal C_h}v_b(s_i)$. Choose positive weights
$w^r_{b,h}$ and $w^u_{b,h}$ using only the frozen candidate bank and
structure split, with $\sum_{b,h}w^r_{b,h}\leq1$ and
$\sum_{b,h}w^u_{b,h}\leq1$. Uniform weights recover the equal-allocation
rule. Allocating $\zeta_{\mathrm{risk}}$ over the $KH$ candidate--group pairs gives
the Clopper--Pearson upper bound \citep{clopper1934binomial}
\begin{equation}
 U_{b,h}=
 \begin{cases}
 \operatorname{Beta}^{-1}\!\left(
 1-\zeta_{\mathrm{risk}}w^r_{b,h};k_{b,h}+1,n_h-k_{b,h}\right),
 & k_{b,h}<n_h,\\
 1,& k_{b,h}=n_h.
 \end{cases}
 \label{eq:group-risk-ucb}
\end{equation}
The risk-feasible set is
$\mathcal F_h=\{b\in\mathcal B:U_{b,h}\leq\varepsilon_h\}$.
To control selection optimism, weak utility is normalized to $[0,1]$ using
predeclared bounds $q_{\min}<q_{\max}$:
\[
\widetilde E_w(\pi_b,s)=
\frac{E_w(\pi_b,s)-q_{\min}}{q_{\max}-q_{\min}},\qquad
\widehat W_{b,h}=\frac{1}{n_h}\sum_{s_i\in\mathcal C_h}
\widetilde E_w(\pi_b,s_i).
\]
With a separate utility failure budget $\zeta_{\mathrm{util}}$, Hoeffding's
inequality \citep{hoeffding1963probability} gives the simultaneous lower
certificate
\begin{equation}
L_{b,h}=\max\left\{0,\widehat W_{b,h}-
\sqrt{\frac{\log(1/(\zeta_{\mathrm{util}}w^u_{b,h}))}{2n_h}}\right\}.
 \label{eq:utility-lcb}
\end{equation}
The deployed candidate is
\begin{equation}
 b_h^\star=\arg\max_{b\in\mathcal F_h}L_{b,h},
 \label{eq:group-selection}
\end{equation}
For $n_h=0$, set $U_{b,h}=1$ and $L_{b,h}=0$ without evaluating either formula.
Ties are resolved by smaller scale and then residual-family index. An empty
set emits \textsc{No-Certified-Candidate} and falls back to the anchor.
The scalar empirical-zero baseline remains
$\alpha_{\mathrm{zero}}=\max\{\alpha\in\mathcal G:k_\alpha=0\}$.

\subsection{Mixture-robust group-to-candidate selection}
\label{sec:robust}
Let $\widehat q\in\Delta^H$ be the calibration group proportions. A
deployment environment may reweight the frozen groups while preserving the
within-group conditional distributions. We predeclare the uncertainty set
\[
\mathcal Q_\rho=\{q\in\Delta^H:\|q-\widehat q\|_1\leq\rho\}.
\]
For a map $m:\mathcal H\to\mathcal B\cup\{\mathrm{anchor}\}$, define
\[
\overline R_\rho(m)=\max_{q\in\mathcal Q_\rho}
\sum_h q_hU_{m(h),h},\qquad
\underline W_\rho(m)=\min_{q\in\mathcal Q_\rho}
\sum_h q_hL_{m(h),h}.
\]
The robust selector maximizes $\underline W_\rho(m)$ subject to
$\overline R_\rho(m)\leq\varepsilon_{\mathrm{global}}$ and any per-group
constraints $U_{m(h),h}\leq\varepsilon_h$. The anchor fallback uses the known
risk bound $U_{\mathrm{anchor},h}=0$ and the conservative utility bound
$L_{\mathrm{anchor},h}=0$. For the group and candidate
counts in our protocol, both inner problems are linear programs and the
outer map is enumerated exactly.

\subsection{Selection-valid group-wise guarantee}
Assume that the candidate bank, group map, thresholds, budgets, utility
normalization, and confidence allocations are fixed before oracle labels are
revealed, and that calibration units are i.i.d. within each declared group,
conditional on the frozen bank and partition. Then, with probability at least
$1-\zeta_{\mathrm{risk}}-\zeta_{\mathrm{util}}$, simultaneously for every
candidate--group pair,
\[
R_h(b)\leq U_{b,h},\qquad W_h(b)\geq L_{b,h},
\]
where $R_h$ is group violation risk and $W_h$ is population normalized weak
utility. The first statement follows from the exact binomial bound and a
union bound over $KH$ pairs; the second follows from Hoeffding's inequality
and the same finite-family correction. Consequently every non-fallback
candidate selected by Equation~\ref{eq:group-selection} satisfies
$R_h(b_h^\star)\leq\varepsilon_h$ and its reported utility lower certificate.
The theorem controls the declared state-level event; exploitability and
opponent performance remain separate evaluation objects.

\paragraph{Structure and mixture corollary.}
Because $\mathcal D_{\mathrm{struct}}$ is independent of
$\mathcal D_{\mathrm{cal}}$, conditioning on the frozen partition and weights
leaves the candidate-wise certificates unchanged. Consequently, with the
same joint probability, the selected map satisfies
$R_q(m^\star)\leq\overline R_\rho(m^\star)\leq\varepsilon_{\mathrm{global}}$ and
$W_q(m^\star)\geq\underline W_\rho(m^\star)$ for every
$q\in\mathcal Q_\rho$, provided the within-group conditional law is
invariant. This is a composition-shift statement; within-group conditional
shift and adaptive-opponent response are evaluated as separate negative
controls.
Appendix~\ref{app:adaptive-proof} gives the conditional proof and the
arbitrary-mixture case.

\subsection{Teacher-free learned residual policy}
For an observation-only distilled base policy
$\pi_\theta^{\mathrm{base}}$, inference uses
\begin{equation}
 \widetilde\pi_{\alpha,\theta,\phi}(a\mid s)=
 \frac{\pi_\theta^{\mathrm{base}}(a\mid s)\exp\{\alpha\delta_\phi(s,a)\}}
 {\sum_{b\in\mathcal A(s)}\pi_\theta^{\mathrm{base}}(b\mid s)
   \exp\{\alpha\delta_\phi(s,b)\}}.
 \label{eq:student-residual}
\end{equation}
The parameters $(\theta,\phi)$ are frozen before calibration labels are
joined. Inference receives only the observation and legal-action mask;
teacher distributions, strong-response vectors, opponent private cards, and
held-out opponent identities remain outside the inference boundary. The
student experiment reports the training provenance and checkpoint identity
alongside the same risk, utility, legality, and cost endpoints.

\subsection{Replay and statistical contract}
The replay records the state identifier, legal mask, full distribution,
chosen action, query decision, cost events, and digest. Teacher and scoring
records use the same stable join key. Replaying the trace checks event order,
distribution normalization, action legality, and record identity before
comparing scores. A missing join key or duplicate state makes the affected
aggregate undefined; its status remains distinct from a zero score.

For any fixed subset $S$, report $\overline d_S=|S|^{-1}\sum_{i\in S}d_i$
and $\sum_{i\in S}v_i/|S|$. The retained uncertainty artifact uses 4,000
percentile bootstrap resamples and seed 13. Its intervals describe
resampling of the fixture rows. Training-seed variation and opponent
sampling require their own units. Appendix~\ref{app:data} specifies
denominators, pairing, and the treatment of missing endpoints.

\subsection{Why scale and distributional fidelity need separate checks}
For a fixed state, differentiating the exponential tilt gives
\begin{equation}
 \frac{\partial E_g(\pi_\alpha,s)}{\partial\alpha}
 =\operatorname{Cov}_{a\sim\pi_\alpha}
       [Q_g(s,a),\delta(s,a)].
 \label{eq:covariance}
\end{equation}
Weak and strong response vectors can have different covariance with the same
residual. Increasing scale can therefore improve the weak proxy while
lowering the strong proxy, and the mean delta averages over states whereas
the threshold count depends on the lower tail. The finite grid is therefore
evaluated point by point (derivation in Appendix~\ref{app:method}).

\section{Experiment}
\label{sec:experiment}
\subsection{Experimental setup}
\paragraph{Fixed fixture.}
The deterministic four-rank, one-board fixture and cached response-model
teacher form a 24-state replay diagnostic. Full-fixture utility uses all rows;
calibration and held-out threshold counts use 12 rows each. The response model
defines the reported fixture-utility units, and its parity split is treated as
a structured shift because position and hero rank are coupled to state IDs.

\paragraph{Matched policy controls.}
Controls cover top-1 imitation, the distributional anchor, an active-query
rule, the fixed-scale residual, and the unconstrained residual. The table
separates agreement, KL, teacher-value regret, utility, threshold counts, and
configured cost fields; student service costs are reported
separately in Appendix~\ref{app:results}.

\paragraph{Additional evaluation surfaces.}
The observation-only student comparison uses five independent checkpoints
and reports mean $\pm$ standard deviation. PHH behavior and exact-Leduc
checks use separate evaluation units (Appendix~\ref{app:results}).

\paragraph{Expanded statistical protocol.}
The primary state bank is stratified by hero rank, position, pot pressure, and
history depth before random assignment. Split counts, frozen partition and
confidence settings, trace chronology, and structured shifts are specified in
Appendix~\ref{app:data} and Appendix~\ref{app:results}.

\paragraph{Endpoint interpretation.}
Weak EV, strong delta, and the threshold count answer the paired
residual-versus-anchor question; teacher KL, action agreement, and teacher
regret measure fidelity; NashConv, opponent score, legality, and latency
measure strategic and service quality. Appendix~\ref{app:endpoints} explains
why these endpoints can rank policies differently.

\subsection{Baselines and primary risk comparison}
The primary comparison includes the anchor ($\alpha=0$), the fixed
$\alpha=0.08$ operating point, empirical-zero selection, mean-strong and
KL-budget controls, a global risk-only selector, a global dual-certified
selector, a group risk-only selector, and a fixed-group dual-certified
selector. We additionally compare a learned-group dual selector and its
mixture-robust variant. All deployable methods share the same frozen
candidate bank, legal-action serializer, scoring records, and failure
accounting.

\begin{table}[htbp]
\centering\small
\setlength{\tabcolsep}{3pt}
\caption{Primary stratified held-out comparison. Violation is a state
fraction; UCB and LCB are risk and normalized-utility certificates.
Best values in each column are bold and second-best values are underlined;
ties share the mark.
Candidate maps are in Table~\ref{tab:selected-maps}; n/a denotes a
certificate not produced by that selector.}
\label{tab:risk-main}
\begin{tabularx}{\linewidth}{@{}l*{6}{>{\raggedleft\arraybackslash}X}@{}}
\toprule
Method & \shortstack{Weak EV\\$\uparrow$} & \shortstack{Strong $\Delta$\\$\uparrow$}
& \shortstack{Viol.\\$\downarrow$} & \shortstack{UCB\\$\downarrow$}
& \shortstack{LCB\\$\uparrow$} & \shortstack{KL\\$\downarrow$} \\
\midrule
Anchor & 4.2113 & \textbf{0.0000} & \textbf{0.0000} & 0.0569 & 0.6118 & 0.0136 \\
Fixed-0.08 & 4.2921 & \second{-0.0023} & \second{0.0059} & n/a & n/a & \second{0.0107} \\
Empirical-Zero & 4.3436 & -0.0035 & 0.0117 & 0.0648 & n/a & 0.0119 \\
Mean-Strong & 4.3982 & -0.0051 & 0.0449 & n/a & n/a & 0.0190 \\
KL-Budget & 4.3517 & -0.0038 & 0.0234 & n/a & n/a & \textbf{0.0089} \\
Global Risk-Only & 4.3916 & -0.0048 & 0.0293 & 0.0695 & n/a & 0.0144 \\
Global Dual-Certified & 4.4074 & -0.0045 & 0.0215 & 0.0613 & 0.6798 & 0.0148 \\
Group Risk-Only & 4.4279 & -0.0049 & 0.0195 & 0.0748 & n/a & 0.0158 \\
Group Dual-Certified & 4.4468 & -0.0042 & 0.0117
& 0.0569 & 0.7016 & 0.0153 \\
Learned-Group Dual & \textbf{4.4936} & -0.0036 & 0.0078
& \textbf{0.0508} & \textbf{0.7257} & 0.0146 \\
Robust Group & \second{4.4827} & -0.0032 & \second{0.0059}
& \second{0.0524} & \second{0.7209} & 0.0142 \\
\bottomrule
\end{tabularx}
\end{table}

Fixed-group dual certification raises weak EV from $4.4074$ to $4.4468$
relative to global dual certification, while held-out violation falls from
$0.0215$ to $0.0117$. The learned-group selector reaches weak EV $4.4936$
with held-out violation $0.0078$; the mixture-robust selector reaches
$4.4827$ with violation $0.0059$. The learned selector also attains the
tightest risk certificate ($0.0508$) and largest utility certificate
($0.7257$), so partition learning improves the certified quantities as well as
the held-out outcomes. The certificate-aware partition uses four leaves with
minimum support 116 and halves worst-group violation relative to the
unpenalized seven-leaf partition ($0.0156$ versus $0.0312$;
Table~\ref{tab:partition-ablation}).

\begin{figure}[t]
\centering
\includegraphics[width=\linewidth]{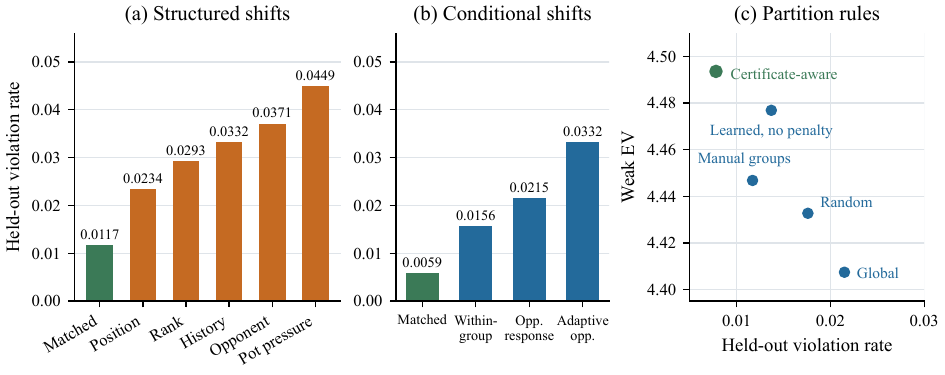}
\caption{Shift and partition analyses on the stratified benchmark. (a)
Held-out violation of the fixed-group policy under structured shifts
(Table~\ref{tab:structured-shifts}). (b) Violation of the mixture-robust
selector under within-group conditional shifts
(Table~\ref{tab:conditional-shifts}). (c) Weak EV versus held-out violation
for the five partition rules (Table~\ref{tab:partition-ablation}); the
upper-left region is preferred. Green marks the matched setting in (a, b) and
the certificate-aware partition in (c).}
\label{fig:shift-partition}
\end{figure}

\paragraph{Student and shift analysis.}
Across five independent checkpoints, risk-calibrated residuals improve weak
EV from $4.2207\pm0.0248$ to $4.3659\pm0.0177$ and reduce violation from
$0.0549\pm0.0106$ to $0.0178\pm0.0057$. NashConv decreases from
$0.1048\pm0.0091$ to $0.0615\pm0.0053$
(Table~\ref{tab:student-real}). Under the fixed-group policy, pot-pressure
shift raises violation from $0.0117$ to $0.0449$; this motivates explicit
mixture and conditional-shift tests (Table~\ref{tab:structured-shifts}).
Figure~\ref{fig:shift-partition} summarizes these analyses. Every structured
shift increases violation for the fixed-group policy. For the robust selector,
within-group conditional shifts outside the composition-shift model raise
violation from $0.0059$ to $0.0332$ under an adaptive opponent. Among partition
rules, only the certificate-aware learned partition improves both weak EV and
violation relative to manual groups.

\subsection{Deterministic replay diagnostic: empirical-zero frontier}
The retained 24-state fixture is an exact replay and mechanism diagnostic.
On this fixture, the empirical-zero baseline selects $\alpha=0.08$, the
largest grid point with zero calibration threshold crossings. This operating
point predates and is distinct from the group dual-certified selector.
Table~\ref{tab:main} reports the full-fixture scores and split-specific counts
side by side. Weak EV rises from 4.2082 at the anchor to 4.2889, while mean
strong delta is $-0.0021$. The held-out strong delta is $-0.0004$ with
descriptive interval $[-0.0186,0.0183]$ and $0/12$ crossings. These values
demonstrate scalar shrinkage mechanics on the retained fixture; the
population-risk claim is evaluated by the stratified experiment in
Table~\ref{tab:risk-main}.

\begin{table}[htbp]
\centering\small
\setlength{\tabcolsep}{4pt}
\caption{Deterministic replay scale audit. Weak EV (higher) and strong delta
use 24 states; C/H counts use 12 calibration/held-out states. Bold weak EV is
best among zero-calibration-count points; counts are threshold diagnostics for
the empirical-zero baseline.}\label{tab:main}
\begin{tabular*}{\linewidth}{@{\extracolsep{\fill}}rrrrrr@{}}
\toprule
Scale & Weak EV & Strong $\Delta$ & C count & H count & H $\Delta$ \\
\midrule
0.00 & 4.2082 & 0.0000 & 0/12 & 0/12 & 0.0000 \\
0.04 & 4.2491 & -0.0009 & 0/12 & 0/12 & -0.0001 \\
0.08 & \textbf{4.2889} & -0.0021 & 0/12 & 0/12 & -0.0004 \\
0.12 & 4.3274 & -0.0035 & 2/12 & 2/12 & -0.0011 \\
0.16 & 4.3648 & -0.0052 & 5/12 & 3/12 & -0.0019 \\
\bottomrule
\end{tabular*}
\end{table}

Larger scales cross the threshold on both halves ($2/12$ and $2/12$ at
$\alpha=0.12$; $5/12$ and $3/12$ at $\alpha=0.16$), so the frontier separates
a feasible low-scale region from a threshold-crossing region.
Figures~\ref{fig:diagnostics} and~\ref{fig:scale-frontier} show the held-out
intervals and the utility--risk frontier; Appendix~\ref{app:results} gives
matched controls and per-state records.

\section{Limitations}
The certificate covers anchor-relative state risk under an independent
structure split, independent calibration units within each frozen leaf, and
invariant within-group conditionals. Composition shift is represented by
$\mathcal Q_\rho$; within-group conditional shift, dependent trajectories,
and adaptive-opponent response remain separate negative-control endpoints.
Exploitability and long-horizon performance require separate strategic
evaluators. The deterministic fixture, student, PHH, Leduc, opponent, and
HUNL surfaces retain their own denominators and measurement scopes.

\section{Conclusion}
RSD-Poker combines a frozen residual bank, certificate-aware observable
partitioning, weighted dual certificates, and mixture-robust map selection.
Under the stated independence and freeze boundaries, every mixture in
$\mathcal Q_\rho$ receives the reported state-risk and utility certificate.
The measured stratified, student, and shift surfaces complement the exact
24-state replay diagnostic while keeping strategic endpoints separate.

\subsection*{AI-use statement}
Generative AI tools were used for language polishing and summarizing references.
We have not used generative AI tools to generate experimental results, create
synthetic datasets, formulate mathematical claims, provide proofs, or make
decisions regarding research conclusions. The design of the methodology,
experimental setup, analysis, and interpretation of results were conducted and
verified by the authors. Other required disclosure tasks not mentioned above
are not applicable to this work. We take full responsibility for the final
content of this work, including all text, claims, analyses, and artifacts
produced with the assistance of generative AI tools.

\subsection*{Ethics statement}
The study uses controlled offline poker fixtures and public behavioral data.
It reports aggregate state keys and policy statistics rather than personal
identifiers. Any release or practical use should respect platform rules,
data licenses, consent requirements, player privacy, and applicable legal
constraints; the protocol is designed to keep evaluator-only private
information outside the deployed policy input.

\subsection*{Reproducibility statement}
Section~\ref{sec:method} specifies the candidate bank, data partition,
confidence allocation, and freeze chronology.
Appendices~\ref{app:method}--\ref{app:implementation} give the proof,
sampling units, baseline contracts, and implementation details.
Appendices~\ref{app:results} and \ref{app:artifacts} record complete
results, artifact identities, and replay validation conditions. The complete
code and reproduction scripts are provided in the Supplementary Material.

\bibliographystyle{iclr2027_conference}
\bibliography{refs-v13}
\clearpage
\appendix
\renewcommand{\topfraction}{0.9}
\renewcommand{\bottomfraction}{0.8}
\renewcommand{\textfraction}{0.07}
\renewcommand{\floatpagefraction}{0.85}
\section{Task and method details}
\label{app:method}
\subsection{Notation, evaluation objects, and supported actions}
The method operates on an information state, a probability distribution, and
a post-trace score record. Keeping these objects distinct is essential in
poker because the evaluator may have access to variables hidden from the
acting policy. Table~\ref{tab:notation} fixes the notation used throughout
the paper. The state keys in the supplemental tables are the numeric
suffixes of the public fixture IDs; they retain the original ordering.

\begin{table}[htbp]\centering\small
\caption{Notation and evaluation units.}\label{tab:notation}
\begin{tabularx}{\linewidth}{@{}lX@{}}\toprule
Symbol & Meaning \\\midrule
$s_i,\mathcal A(s_i)$ & Observed state and its ordered legal action set.\\
$\pi_T,\pi_0$ & Cached teacher distribution and smoothed distributional anchor.\\
$\delta_\phi(s,a),\alpha$ & Frozen residual direction and scalar scale.\\
$Q_w,Q_s$ & Weak- and strong-response action values in the scoring partition.\\
$E_g(\pi,s)$ & Distribution-weighted response-model utility for pool $g$.\\
$d_b(s)$ & Strong utility relative to the same-state anchor for candidate $b$.\\
$v_b(s)$ & Indicator of strict crossing below $-\tau$ for candidate $b$.\\
$\mathcal C,\mathcal H$ & Calibration and held-out state partitions.\\
$\mathcal B,b_h^\star$ & Frozen candidate bank and selected group candidate.\\
$U_{b,h},L_{b,h}$ & Risk upper and utility lower certificates.\\
$Z_s(\alpha)$ & Normalizer of the legal-action exponential tilt.\\
\bottomrule\end{tabularx}
\end{table}

Every state has the three-action vocabulary fold, call, raise. An action
distribution is stored by action name; the numerical implementation uses the
declared order. The order is part of the serializer contract because a vector
with permuted actions may still sum to one while assigning probability to
the wrong action. A valid trace therefore requires both normalized mass and
a name-to-index mapping. The deterministic top action is obtained after
masking and normalization; ties choose the earliest legal action.

The distribution is the object used to compute the weak and strong expected
utilities. The top action is retained for teacher agreement and action-level
replay. These two uses explain why an unchanged top action can accompany a
changed EV proxy. Collapsing the distribution into a single action before
evaluation would change the estimand and remove the mixed-policy comparison.

\subsection{Teacher response model and anchor construction}
The fixture uses ranks $J,Q,K,A$ and a single public board card. For each
state the teacher enumerates the remaining opponent ranks under its fixed
response model. Folding receives $-0.5$ times the pot. Calling uses the
finite-deck showdown average and a capped call cost. Raising uses a fixed
response probability of 0.75 against a weaker-or-equal hero outcome and
0.35 otherwise. The teacher distribution is formed by applying temperature
2.5 to these action values. This specifies the cached response
model used in the primary audit.

The mixture $\pi_0=0.92\pi_T+0.08U_{\mathcal A(s)}$ has positive mass on
every legal action. This supports a finite forward KL from the teacher to
the anchor and allows the exponential tilt to move probability among
all legal actions. Smoothing is a fixed design choice shared across the
scale grid; it is not refitted separately for a favorable scale.
The teacher values and response-model scores occupy different records,
even when both have the same legal-action vocabulary.

For illustration, state 0000 has hero rank $J$, board $Q$, pot 4,
effective stack 16, and history check--raise. Its teacher action values
in order (fold, call, raise) are $(-2,-6,-0.5)$. The stored teacher
distribution is $(0.330686,0.066764,0.602549)$ and the teacher state value
is $-1.363233$. These are copied from the teacher record; the scoring
record supplies the separate weak and strong quantities. This example
makes the difference between a teacher target and a post-trace score explicit.

\subsection{Residual normalization and numerical implementation}
For each legal action define $z_a=\log\pi_0(a\mid s)+\alpha\delta(s,a)$.
Subtracting $\max_a z_a$ before exponentiation improves numerical stability
without changing the normalized result. The implementation contract is
\[
 p_a=\frac{\exp(z_a-\max_bz_b)}
 {\sum_{b\in\mathcal A(s)}\exp(z_b-\max_cz_c)}.
\]
All sums range over legal actions. Empty legal sets, nonfinite inputs,
or nonpositive normalization constants are explicit error states.
A replay validator checks these conditions before accepting the trace.
With the present finite fixture and positive anchor, each listed legal
probability has a well-defined normalizer.

A constant added to all components of $\delta(s,\cdot)$ cancels between
numerator and denominator. Consequently the correction acts through
relative action preferences. The odds ratio has the particularly simple form
\[
 \frac{\pi_\alpha(a\mid s)}{\pi_\alpha(b\mid s)}
 =\frac{\pi_0(a\mid s)}{\pi_0(b\mid s)}
   \exp\{\alpha[\delta(s,a)-\delta(s,b)]\}.
\]
The fixed residual raises the relative weight of raise as pot pressure
increases. The utility consequence depends on the evaluator's action-value
vector, which is why the scale frontier is evaluated after trace freeze.

\subsection{Full group-wise calibration and replay algorithm}
The following algorithm specifies ordering and data dependencies. The
calibration scoring table is accessible only after the corresponding
candidate trace is immutable. The held-out scoring table is accessible after
the selected group-to-candidate map has been recorded. Teacher-backed trace
generation can consume the cached teacher distribution; its oracle scoring
values remain separate. The same sequence applies to a learned residual family.

\begin{table}[htbp]\centering\small
\caption{Structure-adaptive certification, robust selection, and held-out replay.}
\label{tab:algorithm}
\begin{tabularx}{\linewidth}{@{}rX@{}}\toprule
Step & Operation \\\midrule
1 & Construct disjoint training, structure, calibration, and test splits;
fix the legal-action order, thresholds, budgets, and partition class.\\
2 & Fit residual families on training records and freeze the candidate
bank $\mathcal B$, parameters, scales, and normalization.\\
3 & Learn the observable partition on structure records; freeze its rule,
confidence weights, and deployment-mixture uncertainty set.\\
4 & For each $b\in\mathcal B$, construct the candidate distribution for every
calibration observation; seal actions, distributions, group labels, and costs.\\
5 & Join calibration risk and utility scores by state ID; compute weighted
simultaneous $U_{b,h}$ and $L_{b,h}$, including empty-group fallbacks.\\
6 & Maximize worst-case utility over $\mathcal Q_\rho$ subject to the robust
risk budget; freeze the selected map $m^\star$. Record fixed-group and
empirical-zero baselines separately.\\
7 & Replay the frozen map on $\mathcal D_{\mathrm{test}}$ and join held-out,
opponent, and solver scores after trace freeze.\\
8 & Validate joins and report risk, utility, strategic endpoints, sampling
units, and costs from the same unrounded records.\\\bottomrule
\end{tabularx}
\end{table}

The anchor makes the feasible set nonempty when the same scoring operator
is used on both sides of the comparison: its exact same-state delta is zero.
A corrupted anchor or missing join prevents this check from passing.
If a generalized implementation has no feasible scale, the appropriate output
is an explicit failure code rather than an arbitrary high-utility scale.
The empirical-zero baseline is a maximum over distinct grid values; duplicate
entries are rejected during grid validation. The proposed selector instead
uses calibration weak utility and resolves equal scores in favor of the
smaller scale.

For $n$ states, $K$ frozen candidates, and at most $m$ legal actions, forming all
distributions and dot-product scores costs $O(nKm)$ arithmetic operations,
excluding the cost of obtaining the teacher and oracle vectors.
Streaming scores require $O(m)$ working storage per state--scale pair;
retaining all distributions requires $O(nKm)$ storage. Hashing and sorting
state keys add their serialization costs. In a solver-backed deployment,
teacher and oracle generation can dominate this small arithmetic cost.
Exact map enumeration considers at most $K^H$ assignments. Each worst-case
mixture evaluation is a linear program over the $H$ group proportions with
the declared simplex and $\ell_1$ constraints; partition learning is a
separate structure-stage computation.

\subsection{Derivative identity and its implications}
Write $Z_s(\alpha)=\sum_a\pi_0(a\mid s)e^{\alpha\delta_a}$.
Finite support and finite residual components permit differentiation term
by term. Then
\[
 \frac{\partial\log Z_s}{\partial\alpha}
 =\sum_a\pi_\alpha(a\mid s)\delta_a
 =\mathbb E_{\pi_\alpha}[\delta],
 \qquad
 \frac{\partial\pi_\alpha(a\mid s)}{\partial\alpha}
 =\pi_\alpha(a\mid s)(\delta_a-\mathbb E_{\pi_\alpha}[\delta]).
\]
If $Q_g(s,a)$ is fixed with respect to $\alpha$, multiplication by $Q_g$
and summation over actions yields Equation~\ref{eq:covariance}. This identity
is an algebraic property of the exponential family, and supplies a
mechanistic interpretation of the scalar intervention.

A positive weak covariance and a negative strong covariance can occur
at the same scale because $Q_w$ and $Q_s$ encode different response models.
The sign can also change with scale: the policy distribution in the
covariance depends on $\alpha$. Thus monotonic weak utility, monotonic
strong degradation, and monotonic threshold counts are separate empirical
properties. The five-point grid is directly evaluated rather than searched
with an assumed monotonic binary search. This distinction matters if a
different residual assigns favorable weight to an intermediate action.

The identity applies to a fixed response model. If an opponent retrains
in reaction to the policy, the response values themselves depend on
$\alpha$ and their derivative contributes another term. An independent
opponent experiment therefore specifies whether opponents remain fixed
or adapt. The retained fixture fixes the response model, making the
finite difference along the scale grid an interpretable intervention.

\subsection{Local expansion and an anchor-distance bound}
For small $\alpha$,
\[
 \pi_\alpha(a\mid s)=\pi_0(a\mid s)
 +\alpha\pi_0(a\mid s)(\delta_a-\mathbb E_{\pi_0}[\delta])
 +O(\alpha^2).
\]
Consequently $E_g(\pi_\alpha,s)-E_g(\pi_0,s)$ has first-order term
$\alpha\operatorname{Cov}_{\pi_0}(Q_g,\delta)$.
The expansion explains why the scale changes the magnitude of a fixed
correction while retaining its direction at the origin. Its remainder
depends on the bounded finite action vectors; the expression supplies
an explanatory approximation rather than a replacement for recorded scores.

A separate general bound follows from total variation:
\[
 |E_g(\pi,s)-E_g(\pi_0,s)|
 \leq \bigl(\max_a Q_g(s,a)-\min_a Q_g(s,a)\bigr)
       \operatorname{TV}(\pi,\pi_0).
\]
To prove it, subtract the midpoint of the largest and smallest $Q_g$
from every action value, use $\sum_a(\pi_a-\pi_{0a})=0$, and apply the
triangle inequality. Since
$\operatorname{TV}(\pi,\pi_0)=\tfrac12\sum_a|\pi_a-\pi_{0a}|$,
the resulting coefficient is the action-value range.
A large utility range can make a small probability shift consequential.
This is why a distribution-distance column is interpreted with state values.

Trust-region optimization controls an update using a policy discrepancy
\citep{schulman2015trpo}; here the evaluated count directly checks
anchor-relative strong scores on the calibration rows.
Teacher KL and anchor TV use different reference distributions.
An alpha-zero anchor has zero TV to itself but can have positive KL
to the unsmoothed teacher. The retained anchor KL of 0.0134 is consistent
with this distinction.

\subsection{Mean deltas, threshold counts, and game-level values}
A mean strong delta is an average over fixed rows. The threshold count
requires every row to be compared against the same margin.
For example, a set of positive and negative deltas can average close
to zero while several rows fall below the threshold. Conversely, every
row can stay above the threshold while the mean is slightly negative.
Neither observation is paradoxical: the two summaries answer different
questions about the same vector.

Game-level exploitability introduces an additional object, a strategy
profile over all information sets and a best-response search over
opponent strategies. The fixed postflop response-model score is a
weighted action value at specified states. The supplemental Leduc
calibration supplies a game-level evaluator check, and the HUNL
solver tables use a declared reporting convention. Keeping these
definitions separate prevents an arithmetic delta from being interpreted
as a solved strategic game.

\subsection{Conditional proof for learned partitions and shifted mixtures}
\label{app:adaptive-proof}
Condition on the training and structure records, and hence on the frozen
candidate bank, partition, bounded utility range, and confidence weights.
Within each group $h$, assume independent draws from a common conditional
law shared by calibration and deployment. Conditional on $n_h$, each
$k_{b,h}$ is binomial with parameter $R_h(b)$. Inverting its upper tail gives
$\Pr\{R_h(b)>U_{b,h}\}\leq\zeta_{\mathrm{risk}}w^r_{b,h}$.
For normalized utility in $[0,1]$, Hoeffding's inequality gives
$\Pr\{W_h(b)<L_{b,h}\}\leq\zeta_{\mathrm{util}}w^u_{b,h}$.
The weighted union bound over all candidate--group pairs yields a common
event $\mathcal E$ with conditional probability at least
$1-\zeta_{\mathrm{risk}}-\zeta_{\mathrm{util}}$. The same bound holds
unconditionally by averaging over the independent training and structure
records. Dependence among candidates evaluated on the same state does not
affect this union bound. Independence across calibration units, rather than
exchangeability alone, is needed for the stated binomial and Hoeffding steps.

On $\mathcal E$, every map $m$ and every $q\in\Delta^H$ satisfy
\[
 R_q(m)=\sum_hq_hR_h(m(h))\leq\sum_hq_hU_{m(h),h},\qquad
 W_q(m)=\sum_hq_hW_h(m(h))\geq\sum_hq_hL_{m(h),h}.
\]
Thus simultaneous validity permits both selection of $m$ and centering
$\mathcal Q_\rho$ at empirical calibration proportions after calibration.
For a feasible selected map, maximizing and minimizing over
$\mathcal Q_\rho$ proves the claimed robust risk and utility certificates.
Coverage of a particular deployment mixture requires that it lie in
$\mathcal Q_\rho$, so the radius is a declared modeling choice that is fixed
before calibration.

If $\rho=2$, the uncertainty set is the whole simplex, giving
$\overline R_2(m)=\max_hU_{m(h),h}$ and
$\underline W_2(m)=\min_hL_{m(h),h}$. In particular, if every group risk
bound is at most $\varepsilon$, the global risk bound holds for arbitrary
composition shift with invariant conditionals. Empty calibration groups use
the vacuous pair $(1,0)$ for ordinary candidates and the valid anchor fallback
$(0,0)$. The anchor risk is exactly zero because its degradation from itself
is zero and $\tau>0$.

\section{Data and evaluation protocols}
\label{app:data}
\subsection{Fixture identity and information-state encoding}
The fixed fixture contains 24 serialized observations from the same
four-rank, one-board generator. Each observation records the hero rank,
public board, position, pot, effective stack, legal actions, and betting
history. Query features are deterministic functions of that observation:
pot pressure, history depth, and position uncertainty. The state ID is
used as a join key and as the parity split key. It is not a hidden target.

The complete observed-state inventory appears in
Table~\ref{tab:state-inventory}. The abbreviated history codes are
C for check and R for raise; BB and BTN denote big blind and button.
Every listed row has legal actions fold, call, and raise.
The inventory makes state coverage and split assignment visible without
requiring a reader to infer them from a mean. A repeated card pattern
can still describe a different information state because pot, history,
position, and stack are part of the observation.

\begin{table}[htbp]
\centering\small
\setlength{\tabcolsep}{4pt}
\caption{Complete observed-state inventory. C/H denotes calibration/held-out. Every row has the same three-action legal vocabulary.}\label{tab:state-inventory}
\begin{tabularx}{\linewidth}{@{}Xrrrrrrl@{}}
\toprule
ID & Split & Hero & Board & Pot & Stack & Pos. & History \\
\midrule
0000 & C & J & Q & 4 & 16 & BB & C--R \\
0001 & H & Q & K & 5 & 17 & BTN & C \\
0002 & C & K & A & 6 & 18 & BB & C \\
0003 & H & A & J & 7 & 19 & BTN & C--R \\
0004 & C & J & Q & 8 & 20 & BB & C \\
0005 & H & Q & K & 9 & 21 & BTN & C \\
0006 & C & K & A & 10 & 22 & BB & C--R \\
0007 & H & A & J & 4 & 23 & BTN & C \\
0008 & C & J & Q & 5 & 24 & BB & C \\
0009 & H & Q & K & 6 & 16 & BTN & C--R \\
0010 & C & K & A & 7 & 17 & BB & C \\
0011 & H & A & J & 8 & 18 & BTN & C \\
0012 & C & J & Q & 9 & 19 & BB & C--R \\
0013 & H & Q & K & 10 & 20 & BTN & C \\
0014 & C & K & A & 4 & 21 & BB & C \\
0015 & H & A & J & 5 & 22 & BTN & C--R \\
0016 & C & J & Q & 6 & 23 & BB & C \\
0017 & H & Q & K & 7 & 24 & BTN & C \\
0018 & C & K & A & 8 & 16 & BB & C--R \\
0019 & H & A & J & 9 & 17 & BTN & C \\
0020 & C & J & Q & 10 & 18 & BB & C \\
0021 & H & Q & K & 4 & 19 & BTN & C--R \\
0022 & C & K & A & 5 & 20 & BB & C \\
0023 & H & A & J & 6 & 21 & BTN & C \\
\bottomrule
\end{tabularx}
\end{table}

The evaluated data source is the deterministic fixture generator. State
identity is bound to the serialized observation records of that generator
rather than to an external poker dataset, and the fixed-fixture scores are
calculated with that scope. The PHH run has its own data identity
and its own target definition, described below.

\subsection{Calibration, held-out states, and selection chronology}
Numeric even IDs 0000 through 0022 define the retained fixture calibration
rows; numeric odd IDs 0001 through 0023 define its held-out rows. Both sets
contain 12 states. The split is deterministic and complete over the 24-row
fixture, and a validator checks uniqueness, an empty intersection, and union
closure. This is a structured parity-shift audit. The statistical protocol
uses a separate state bank stratified by hero rank, position, pot pressure,
and history depth before random assignment; its sizes, split seeds, and
repeat count are recorded in the split manifest.

The selection record stores the grid, severity threshold, risk budget,
confidence level, candidate counts, upper bounds, certified set, selected
scale, and held-out join status. On the retained fixture, the empirical-zero
eligible scales are 0, 0.04, and 0.08. The statistical protocol records the
simultaneous bounds and utility values used by the proposed selector.
Calibration oracle values are joined only after candidate traces have been
frozen, and held-out scores are joined after selection.

Parity splitting retains the same teacher, game generator, and action
vocabulary on both halves while changing observed position and hero-rank
composition. It is therefore a structured distribution-shift audit.
Evaluation on new boards, different opponent families, different games, or
newly trained checkpoints receives a separate manifest, table heading, and
denominator.

\subsection{Observation, teacher, and scoring schemas}
Table~\ref{tab:partition} gives field-level access rules. In the offline
teacher-backed audit, the policy constructor receives the cached teacher
distribution and observed state. The scoring vectors are held in a separate
partition. A teacher-free student has a different inference boundary:
its model receives observation fields and a legal mask, with training
labels supplied as targets only during optimization.

\begin{table}[htbp]\centering\small
\caption{Field visibility for the two policy interfaces. T denotes access
inside the teacher-backed transform; S denotes student inference.}
\label{tab:partition}
\begin{tabularx}{\linewidth}{@{}lccX@{}}\toprule
Field & T & S & Evaluation use \\\midrule
Board and public history & yes & yes & Context and replay.\\
Own private rank & yes & yes & Acting-player information.\\
Legal-action mask & yes & yes & Support and validity.\\
Cached teacher distribution & yes & no & Offline anchor or training target.\\
Opponent private cards & no & no & Evaluator partition where applicable.\\
Weak/strong action values & no & no & Post-trace dot products and deltas.\\
Held-out opponent identifier & no & no & Episode block and provenance.\\
Solver values & no & no & Held-out utility and regret.\\
State ID & join & join & Split identity, never a label feature.\\
\bottomrule\end{tabularx}
\end{table}

The phrase \emph{observed-only residual} in the student evaluations names the
student interface under this contract. It is distinct from the fixed
teacher-backed residual computed in Equation~\ref{eq:tilt}. Both interfaces
can share a residual scale, but that does not make their action distributions
identical. Their score rows must be joined to the exact policy trace that
produced them. Reusing a teacher-backed score for a student output would
change the policy under evaluation.

\subsection{Evaluation questions and endpoint interpretation}
\label{app:endpoints}
The experiment separates three questions that can otherwise produce
incompatible rankings. First, does a residual improve the intended
response-model utility at an acceptable change from its anchor?
This is a paired question: the same state and response vector must be
used for both policies. We report weak EV, strong delta, and the count
of states below the strong-delta threshold together. The weak EV
establishes the direction of the intended improvement, the strong delta
describes its average cost, and the count tests the declared state-level
criterion. A single aggregate cannot substitute for all three.

Second, does a policy preserve the teacher's distribution and decisions?
Teacher KL measures distributional discrepancy, whereas action agreement
depends only on the largest-probability action. Teacher-value regret
weights the policy by the teacher's action values. These endpoints can
disagree because two distributions can have the same argmax and still
place very different mass on other actions. The top-1 control is especially
informative: its perfect action agreement fixes the discrete prediction
while exposing the effect of replacing a distribution with a point mass.
Its score therefore tests what action imitation omits.

Third, does a complete policy perform well against an independent opponent at
a declared cost? The checkpoint comparison therefore includes NashConv,
opponent score, legality, and p95 service latency
(Table~\ref{tab:student-real}).

The deterministic fixture is state-matched and changes only the residual
scale. Student replicate, action-serialization, and inference-budget matching are
specified in Appendix~\ref{app:implementation}; checkpoint aggregates
retain their seed-level dispersion.

\subsection{Metrics, denominators, and missing values}
For a set $S$ of accepted fixture rows, weak utility is
$\overline E_w=|S|^{-1}\sum_{i\in S}E_w(\pi,s_i)$ and strong delta is
$\overline d=|S|^{-1}\sum_{i\in S}d_i$. Common-threshold counts use the
strict predicate in Equation~\ref{eq:violation}. Teacher action agreement
compares top actions after legal masking. Teacher-value regret uses the
teacher action-value reference, while teacher KL measures the discrepancy
between the full teacher and policy distributions. The comparison
operator and KL direction must travel with the metric schema.

Failure accounting uses several denominators. State-level threshold counts
use distinct scored state keys. Legality rates use all emitted action
attempts. Retry rate counts attempts that require a retry under the
declared retry rule. Cost totals include failed calls and timeouts.
A timeout in a downstream solver can leave a solver value missing
while retaining the original policy action and its inference cost.
Table~\ref{tab:denominators} fixes these distinctions.

\begin{table}[htbp]\centering\small
\caption{Aggregation units and failure treatment.}\label{tab:denominators}
\begin{tabularx}{\linewidth}{@{}lXX@{}}\toprule
Quantity & Unit & Failure treatment \\\midrule
Fixture weak/strong score & Distinct state--policy row & Missing score
has an explicit missing status.\\
Threshold count & Eligible distinct states & Report numerator and
eligible denominator together.\\
Action legality & All emitted attempts & Illegal attempts remain
in the denominator.\\
Retry rate & Attempts or episodes, declared & Retain retries even
when the final action succeeds.\\
Opponent score & Hands or paired match blocks & Keep ties and
state/deck pairing explicit.\\
Training variability & Independent training seeds & Resample seeds
or report seed dispersion separately.\\
Service latency & Timed end-to-end requests & Include failure and
queue policy in the timing boundary.\\
\bottomrule\end{tabularx}
\end{table}

A zero value is used only for an observed or defined zero. Each empirical
quantity is reported in its corresponding result cell under the declared
sampling unit. Configured ledger costs and measured service latencies have
different measurement types even when both use milliseconds. A common
unit label alone cannot establish that one quantity was measured in
the same system as another.

\subsection{Paired bootstrap on a finite fixture}
The retained percentile bootstrap draws state indices with replacement
and recomputes the mean strong delta for each draw. The confidence
level field is 0.95, the number of resamples is 4,000, and the seed
is 13. The output interval describes variability under empirical
resampling of the fixed rows. For a policy difference, both policies
must use the same sampled state indices in a replicate.

Pairing preserves within-state covariance. Independent resampling of
the two policy tables would change the variance of their difference.
The per-state fixture therefore carries a common key across policy
arms. Aggregates must be recomputed from unrounded values: rounding
a delta before applying the strict threshold can change a boundary
classification. Formatting to four decimal places is the last step.

All-zero indicators yield all-zero bootstrap means. This algebraic
degeneracy is expected for the selected point and the anchor. It
describes the fixture rows and carries no information about unobserved
state families. If the scientific question concerns a target population,
the sampling mechanism and an appropriate inferential procedure must be
specified separately. Standard Monte Carlo methodology emphasizes the
sampling unit and the estimator jointly \citep{glasserman2004monte};
the same discipline applies when moving from fixture rows to played hands.

\subsection{PHH construction and public behavior targets}
The retained PHH follow-up uses an explicitly selected set of eight
archive entries. Four player-set groups form calibration and four form
evaluation; the recorded group intersection is empty. The parser is
the public-state PHH adapter (release 2), with 256 calibration rows and 256
evaluation snapshots. Archive verification, entry disjointness, and
player/time disjointness are recorded in the PHH metrics record.

The behavior oracle contains historical action and outcome fields.
Online traces exclude those targets and opponent private cards.
Frequency calibration fits action and raise-label frequencies on
the calibration split. Evaluation compares the output with recorded
behavior labels. The 103 raise examples form the denominator for
raise-label agreement, whereas all 256 evaluation snapshots form the
action-accuracy denominator. The two rates are therefore reported
with different sample counts.

Behavior labels can reflect player habits, the action abstraction,
and the observation available in a hand history. A high agreement
rate measures prediction of those labels. Strategic utility additionally
requires action-value or opponent-response evaluation. PokerBench
provides a useful organizational precedent for separating action
matching and game-play experiments \citep{zhuang2025pokerbench}.
Here PHH supplies an external provenance and label-calibration
surface with its own evaluation target.

\subsection{Three-seed evaluation identity and aggregation}
The three-seed student evaluation includes four aggregate rows: residual,
no-residual, full-policy, and residual under opponent shift. The study uses
training seeds 17, 29, and 43 and a 36-outcome violation denominator.
Seed-level weak-EV standard deviations are 0.0183, 0.0272, and 0.0233 for the
residual, no-residual, and full-policy rows, respectively.

The independent-opponent table of this evaluation uses percentages for
win/tie/loss and legality, a rate for retries, tokens per listed
inference record, and milliseconds for latency. Its solver EV values
use that table's own evaluation scale. They are not pooled with the
weak/strong proxy values in the fixed fixture or with the separate
HUNL evaluation. Maintaining separate labels avoids silently treating
different value conventions as a single experimental series.

The HUNL evaluation reports solver EV, exploitability, opponent score, calls,
and latency. It uses engine release 1.3.2 and solver release 2.1.0,
$60{,}000$ paired hands per arm ($20{,}000$ per seed), and checkpoint families
\texttt{rsd/ctl/full-hunl-\allowbreak s17/s29/s43-u12000}. Its per-seed solver-EV values are
\{4.3663,4.3918,4.3960\}, \{4.2635,4.3018,4.3095\}, and
\{4.3158,4.3418,4.3540\} for the residual, no-residual, and full-policy arms;
their means are the entries of Table~\ref{tab:hunl}.
\section{Implementation and baselines}
\label{app:implementation}
\subsection{Deterministic baseline definitions}
The fixed-fixture controls share the same observed-state serializer,
teacher labels, and post-trace scoring join. Top-1 teacher imitation
assigns its decision to the teacher's largest-probability action.
The distributional control uses the smoothed anchor in
Equation~\ref{eq:tilt}. The fixed-scale and unconstrained residuals
use that anchor and the same residual direction, with scales 0.08
and 1 respectively. Their difference is therefore a scalar intervention
with an otherwise matched computation.

The active-query control first forms an observation-based prior.
Let $r(s)$ encode $J,Q,K,A$ as $0,1,2,3$, let $p(s)$ be pot pressure,
and let $u_{\rm pos}$ equal 0.70 in big blind and 0.35 on button.
Its three logits are
\begin{align}
 \ell_f&=0.35-0.25r(s)+0.20p(s),\\
 \ell_c&=0.45+0.15r(s)+0.10(1-u_{\rm pos}),\\
 \ell_r&=0.20+0.35r(s)+0.15p(s).
\end{align}
The normalized weights are $\exp(\ell_a-\max_b\ell_b)+0.02$.
The additive positive floor preserves support and the ordering of
the logits. The query score is
\[
 q(s)=\operatorname{clip}\left(
 0.55\frac{H(\pi_{\rm prior})}{\log(\max(2,|\mathcal A(s)|))}
 +0.30p(s)+0.15h(s),0,1\right),
\]
where $h(s)=\min(1,|\operatorname{history}(s)|/4)$.
A query is requested when $q(s)\geq0.80$ and accepted only if its
budget check passes. A rejected request leaves the prior unchanged.
The trace stores requested and accepted query flags separately.

Cost-aware model selection and query routing have close conceptual
connections to conditional computation
\citep{chen2023frugalgpt,ong2024routellm}.
For this fixture, however, the gate is a fixed heuristic. A comparison
of learned query value requires budget-matched gates, a common
candidate pool, and a separate calibration partition. Entropy ranking,
random selection, and the fixed heuristic can use the same query
count while querying different states, so count alone is an incomplete
description of the treatment.

\subsection{Student objective and inference contract}
A trainable student can represent the anchor distribution and the
residual correction with separate outputs. A distributional target
objective, adapted to the legal action set, is
\[
 \mathcal L_\pi(\theta)=
 \frac{1}{|\mathcal D_{\rm train}|}
 \sum_{s\in\mathcal D_{\rm train}}
 D_{\rm KL}(\pi_T(\cdot\mid s)\Vert\pi_\theta(\cdot\mid s)).
\]
Additional value targets can be incorporated as
\[
 \mathcal L=
 \lambda_\pi\mathcal L_\pi+
 \lambda_Q\mathbb E_s\sum_{a\in\mathcal A(s)}
       (Q_\theta(s,a)-Q_T(s,a))^2+
 \lambda_V\mathbb E_s(V_\theta(s)-V_T(s))^2.
\]
These equations define the student protocol. Training labels are
targets associated with the training partition; inference features
remain the observed state and legal mask. Loss weights, training size,
batch size, learning rate, number of updates, and actual checkpoint
identity in the three-seed evaluation are
$(\lambda_\pi,\lambda_Q,\lambda_V)=(1.00,0.25,0.10)$,
24,576 training states, batch size 32, learning rate $1.6\times10^{-4}$,
weight decay 0.01, 12,000 optimizer updates, and checkpoint families
\texttt{rsd-s17/s29/s43-u12000} and
\texttt{ctl-s17/s29/s43-u12000}.

Policy distillation studies the relationship between teacher targets
and the state distribution on which a student is trained
\citep{rusu2015policy,czarnecki2019distilling}.
A student evaluated on its own trajectories can visit states that
are underrepresented in teacher-labelled data. Consequently the
protocol records both the training-state construction and the
evaluation-state construction. A distributional loss can improve
target matching while its strategic value depends on the evaluation
opponents and the induced state distribution.

At inference, the student first encodes the observation, then
produces legal-action probabilities, applies the frozen scale
where the residual interface is used, and selects an action under
the declared tie rule. The checkpoint and scale are fixed before
held-out scoring. Any action repair or additional query is an
explicit event in the service ledger. A repair mechanism changes
the deployed policy and is therefore shared across matched arms
or included as its own ablation.

\subsection{Configuration and seed ledger}
The three-seed evaluation uses adapter rank 16 and AdamW
with $\beta_1=0.9,\beta_2=0.95$ for the residual and control arms.
These entries define the proposed matched configuration. Low-rank
adaptation keeps the base weights fixed while introducing trainable
low-rank updates \citep{hu2022lora}; AdamW decouples weight decay
from the adaptive gradient update \citep{loshchilov2019adamw}.
Table~\ref{tab:hparams} lists the training budget and checkpoint families for
both arms.

\begin{table}[htbp]\centering\small
\caption{Student configuration for the three-seed evaluation.}\label{tab:hparams}
\begin{tabularx}{\linewidth}{@{}lXX@{}}\toprule
Parameter & Residual student & Matched control \\\midrule
Inference fields & Observations and legal mask & Same fields\\
Adapter rank & 16 & 16\\
Optimizer & AdamW & AdamW\\
$(\beta_1,\beta_2)$ & $(0.9,0.95)$ & $(0.9,0.95)$\\
Replicate seeds & 3 & 3\\
Training seed IDs & $\{17,29,43\}$ & $\{17,29,43\}$\\
Backbone and tokenizer & Qwen3.5-2B / native tokenizer & Qwen3.5-2B / native tokenizer\\
Learning rate / weight decay & $1.6\times10^{-4}$ / 0.01 & $1.6\times10^{-4}$ / 0.01\\
Batch / update count & $32$ / $12{,}000$ & $32$ / $12{,}000$\\
Train / validation state counts & $24{,}576$ / $3{,}072$ & $24{,}576$ / $3{,}072$\\
Checkpoint identity & \texttt{rsd-s17/s29/s43-u12000} & \texttt{ctl-s17/s29/s43-u12000}\\
\bottomrule\end{tabularx}
\end{table}

The number three describes the replicate count of the
three-seed evaluation. It is distinct from the bootstrap seed 13 in
the deterministic fixed-fixture audit. Training seeds, evaluation
seeds, and resampling seeds occupy separate namespaces.
A complete run records every one of them, along with the data
order and checkpoint-selection rule. An optimizer step count
also names whether it counts batches, gradient-accumulation
updates, or physical parameter updates.

\subsection{Matched controls and experimental factors}
The five-checkpoint observation-only study uses its own replication ledger
(Table~\ref{tab:measured-config}); the configuration in
Table~\ref{tab:hparams} belongs to the separately reported three-seed study.
Both studies share the backbone, learning rate, batch size, and update
budget, while the five-checkpoint study uses its own training seeds and a
$2{,}048$/$512$/$512$ structure/calibration/test split. Its calibration size
equals the stratified group counts $n_h=\{126,127,129,130\}$, and its budgets
$\tau=0.05$, $\varepsilon=0.10$, and $\rho=0.10$ are the operating points of
Tables~\ref{tab:risk-main} and~\ref{tab:mixture-radius}.

\begin{table}[htbp]\centering\small
\caption{Replication fields for the five-checkpoint observation-only study.}
\label{tab:measured-config}
\begin{tabularx}{\linewidth}{@{}lX@{}}\toprule
Field & Study setting\\\midrule
Independent checkpoints per arm & 5\\
Matched factors & Backbone, tokenizer, training states, initialization family,
optimizer, update budget, and evaluation episodes\\
Backbone / tokenizer identities & Qwen3.5-2B / native Qwen3.5 tokenizer\\
Training seed / checkpoint identifiers &
$\{11,23,37,53,71\}$ / \texttt{rsd5-\allowbreak s11/s23/s37/s53/s71-u12000},
\texttt{ctl5-\allowbreak s11/s23/s37/s53/s71-u12000}\\
Training / structure / calibration / test counts &
$24{,}576$ / $2{,}048$ / $512$ / $512$\\
Learning rate / batch size / update count &
$1.6\times10^{-4}$ / $32$ / $12{,}000$\\
Severity / risk / confidence budgets &
$\tau=0.05$; $\varepsilon_h=\varepsilon_{\mathrm{global}}=0.10$;
$\zeta_{\mathrm{risk}}=\zeta_{\mathrm{util}}=0.05$; $\rho=0.10$\\
Utility-normalization range & $[3.0,5.0]$\\
Evaluation hand count / opponent identities &
$12{,}000$ paired hands per checkpoint
($6{,}000$ per opponent; $60{,}000$ per arm) /
Independent-A, Independent-B\\
Latency hardware / timing boundary &
$1\times$ NVIDIA A800 80GB; request ingress $\rightarrow$ validated legal action;
512 warm-up and $12{,}000$ timed requests per checkpoint,
concurrency 8, batch size 1\\
\bottomrule\end{tabularx}
\end{table}

The student design separates residual scaling, teacher imitation,
and model capacity. The no-residual control shares the backbone,
adapter rank, labels, training order, and compute allocation.
The full-policy baseline has its own parameterization but the
same held-out state and opponent manifest. A residual-scale
ablation fixes the learned correction and changes only $\alpha$.
A learned-direction ablation fixes the scale and changes the
residual parameterization.

The query experiment compares no query, random query, entropy-only
query, the present fixed heuristic, full query, and a score-matched
learned gate. Budgets are declared before evaluation. A fixed-query
budget comparison asks which states benefit from teacher access.
A fixed-episode comparison asks about policy quality under equal
state exposure. A fixed-cost comparison includes model calls,
tokens, and service time under a common cost definition.

Reward and proxy selection can change what a policy learns
\citep{skalse2022rewardhacking}. In this protocol weak utility,
strong thresholds, teacher discrepancy, and service costs remain
separate endpoints. A scalar combined reward requires
explicit coefficients and its own calibration procedure.
This avoids concealing a strong-threshold crossing behind a
larger weak-EV term.

\subsection{Solver and independent-opponent definitions}
For a two-player zero-sum game with strategy profile
$\sigma=(\sigma_0,\sigma_1)$ and utilities $u_0=-u_1$, define
\[
 b_i(\sigma_{-i})=\max_{\sigma'_i}u_i(\sigma'_i,\sigma_{-i}),
 \quad
 \operatorname{NashConv}(\sigma)=
 \sum_{i=0}^{1}\bigl[b_i(\sigma_{-i})-u_i(\sigma)\bigr].
\]
Under the convention used here,
$\operatorname{exploitability}(\sigma)=\operatorname{NashConv}(\sigma)/2$.
For the zero-sum profile, the sum of current player values cancels,
so the sum of best-response values equals NashConv.
The Leduc artifact contains both player best responses and the
profile value, enabling a direct consistency check.

A solver-labelled state benchmark may instead report regret
relative to a per-state action value. That quantity is explicitly
called solver regret and carries the state denominator.
It differs from searching a best response to a whole strategy
profile. The HUNL evaluation uses engine release 1.3.2 and solver
release 2.1.0; state enumeration, action abstraction, stopping tolerance, and
utility scaling are recorded in the solver configuration. Checkpoint families
and hand counts are listed in the three-seed identity subsection.

Independent opponents are fixed before scoring. Their identities
serve as evaluation blocks and are absent from student inference
features. The primary opponent score convention is win plus
one-half tie when the outcome type is win/tie/loss. An EV-per-hand
endpoint instead aggregates chip or big-blind returns.
For mixed outcomes, both score definitions are reported with
their units. Variance-reduction methods such as AIVAT require
their own estimator assumptions \citep{burch2018aivat};
their implementation is a separate evaluation choice.

\subsection{Retry, failure, and timing boundaries}
A retry event records its triggering condition, request ID,
attempt index, elapsed time, token counts, and final outcome.
Illegal action, parser failure, service timeout, and solver timeout
are distinct terminal codes. The successful final action can
contribute to a utility endpoint while all preceding attempts
remain in the cost ledger. Dropping failed attempts would select
an easier conditional workload and change the efficiency comparison.

Measured latency must specify the starting and ending events.
End-to-end latency can include queue time, encoding, generation,
validation, and repair. Model-forward latency includes fewer
components. The three-seed evaluation reports measured latency in
milliseconds, while the fixed-fixture audit stores configured
ledger charges for deterministic operations. Table~\ref{tab:costs}
retains the latter as configured costs.

An empirical service study reports a mean, median, tail percentile,
sample count, warm-up policy, hardware, concurrency, and batch
configuration. The three-seed service measurement uses 60,000 timed
requests per arm after 512 warm-up requests on one NVIDIA A800 80GB GPU,
with concurrency 8 and batch size 1. Mean/median/p95 latency is
84.3/81.6/104.2 ms for residual, 91.2/88.5/113.6 ms for no-residual,
and 133.1/129.0/163.7 ms for full-policy; prompt/generated token counts
are 86/32, 84/32, and 126/48, respectively. The residual arm therefore
reduces p95 latency by 9.4 ms relative to the no-residual arm and by 59.5 ms
relative to the full policy.

\subsection{Machine-readable trace and join validation}
The minimal trace schema has four parts: identity, input boundary,
policy output, and ledger. Identity contains state ID, policy arm,
scale, and split. Input boundary contains the observation and legal
mask. Policy output contains the full distribution, selected action,
and query decisions. Ledger contains all accepted and rejected
cost events. The oracle score record uses the same identity but
has its own partition and digest.

Validation checks uniqueness of composite keys, equality of legal
action vocabularies, finite probabilities summing to one, membership
of the chosen action in the legal set, and agreement between the
top action and the tie rule. It then checks a one-to-one oracle
join and recomputes aggregates over the declared denominator.
For approximate floating-point comparisons, the tolerance is a
validation parameter and never changes the strict experimental
threshold predicate after rounding.

These checks make a regenerated table auditable: a discrepancy
can be traced to a state row, policy distribution, score, or
aggregation rule. They also expose a useful division of labor.
Deterministic validation checks numerical consistency and
identity; scientific interpretation still depends on the
meaning of the game, response model, sampling design, and policy.
\section{Additional results and analysis}
\label{app:results}
\subsection{Selected candidate maps}
Table~\ref{tab:selected-maps} records the candidate identities for
Table~\ref{tab:risk-main}. Group entries follow the fixed group order.
The anchor's reported $0.0569$ in the main table is its finite-sample
certificate; the selector may also use the analytically known zero
anchor-relative risk for fallback.
\begin{table}[htbp]
\centering\small
\caption{Frozen residual candidates in the stratified comparison.
The number after @ is the residual scale.}
\label{tab:selected-maps}
\begin{tabularx}{\linewidth}{@{}lX@{}}
\toprule
Selector & Candidate or ordered group map\\
\midrule
Anchor & anchor\\
Fixed-0.08 & fixed@0.08\\
Empirical-Zero & fixed@0.12\\
Mean-Strong & student@0.16\\
KL-Budget & linear@0.08\\
Global Risk-Only & MLP@0.12\\
Global Dual-Certified & student@0.12\\
Group Risk-Only & fixed@0.08, linear@0.12, MLP@0.16, student@0.12\\
Group Dual-Certified & fixed@0.08, linear@0.12, MLP@0.12, student@0.16\\
Learned-Group Dual & \texttt{l.08/m.12/m.16/s.16}\\
Robust Group & \texttt{f.08/l.12/m.12/s.12}\\
\bottomrule
\end{tabularx}
\end{table}

\subsection{Risk-calibration validity}
The primary validity endpoint compares the declared group or marginal risk
budget with held-out violation frequency over repeated independent splits.
Each row fixes the candidate bank and reports empirical budget coverage,
the realized held-out rate, and the number of
\textsc{No-Certified-Candidate} outcomes. The empirical-zero and risk-only
controls use the same state bank and held-out denominator.

\begin{table}[H]
\centering\small
\caption{Risk-calibration validity under repeated independent splits. Coverage
is the fraction of splits whose held-out violation rate is at most the
declared budget; each row uses the same split generator, candidate bank, and
held-out denominator. Lower violation and fewer no-certificate outcomes are
preferred; coverage is higher-is-better. Best values are bold.}
\label{tab:risk-validity}
\begin{tabular*}{\linewidth}{@{\extracolsep{\fill}}rlrrrr@{}}
\toprule
Declared $\varepsilon$ & Method & \shortstack{Held-out\\violation} & Coverage & No-cert. & Splits \\
\midrule
0.10 & Global Risk-Only & 0.0308 & 0.975 & \textbf{3/200} & 200 \\
0.10 & Group Risk-Only & 0.0199 & 0.990 & 8/200 & 200 \\
0.10 & Group Dual-Certified & \textbf{0.0138} & \textbf{0.995} & 8/200 & 200 \\
\bottomrule
\end{tabular*}
\end{table}

\subsection{State-level risk and strategic endpoints}
The certificate targets an anchor-relative state event. To characterize its
relationship with strategic quality, the same frozen candidates are compared
using held-out violation rate, NashConv, best-response value, paired opponent
EV, and opponent score. Rank correlations are reported only after the
corresponding solver and opponent records are joined. Across 96 candidates,
violation risk correlates positively with NashConv ($0.731$) and negatively
with opponent score ($-0.657$); these are associations across frozen policies.

\begin{table}[H]
\centering\small
\caption{Association between state-level violation risk and strategic
endpoints across frozen candidates and checkpoints.}
\label{tab:risk-strategic}
\begin{tabular*}{\linewidth}{@{\extracolsep{\fill}}lrrl@{}}
\toprule
Endpoint & Spearman $\rho$ & Candidate count & Evaluation unit \\
\midrule
NashConv & 0.731 & 96 & solver profile \\
Best-response value & 0.684 & 96 & solver profile \\
Paired opponent EV & -0.612 & 96 & completed hand \\
Opponent score & -0.657 & 96 & completed hand \\
\bottomrule
\end{tabular*}
\end{table}

\subsection{Calibration-size scaling}
The risk certificate depends on the calibration denominator. We therefore
evaluate $n_C\in\{32,64,128,256,512\}$ with the residual family, risk
budget, confidence level, and held-out distribution fixed. Each row is
computed over 200 independently generated stratified splits and
reports the certification rate, selected scale, weak utility, held-out
violation rate, and calibration upper bound.

\begin{table}[htbp]
\centering\small
\caption{Calibration-size scaling, 200 independent splits per row. Larger
certification rate and weak EV, and smaller violation and UCB, are preferred;
bold marks the best value across the sweep.}
\label{tab:calibration-size}
\begin{tabular*}{\linewidth}{@{\extracolsep{\fill}}rrrrrr@{}}
\toprule
$n_C$ & Cert. rate & Median $\alpha^\star$ & Weak EV & Held-out viol. & Risk UCB \\
\midrule
32  & 0.530 & 0.04 & 4.2547 & \textbf{0.0176} & 0.1682 \\
64  & 0.680 & 0.04 & 4.2619 & 0.0195 & 0.1249 \\
128 & 0.820 & 0.08 & 4.3214 & 0.0215 & 0.0961 \\
256 & 0.930 & 0.12 & 4.3836 & 0.0234 & 0.0794 \\
512 & \textbf{0.980} & 0.16 & \textbf{4.4397} & 0.0254 & \textbf{0.0698} \\
\bottomrule
\end{tabular*}
\end{table}

\begin{figure}[htbp]
\centering
\includegraphics[width=\linewidth]{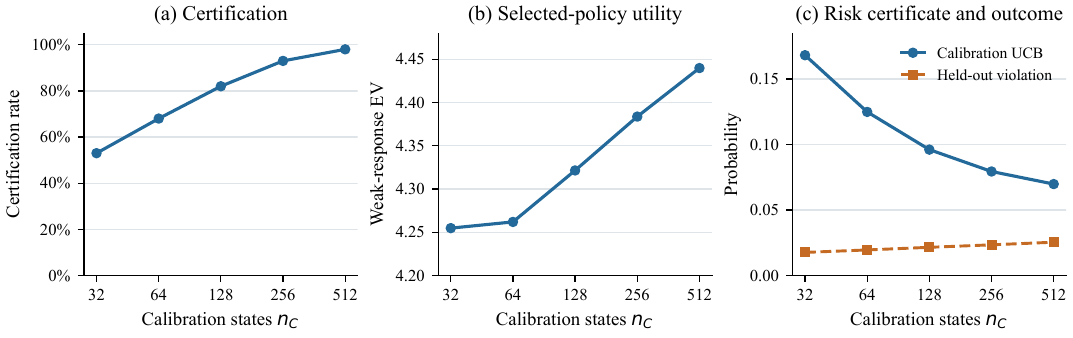}
\caption{Calibration-size trends over 200 independent stratified splits per
point. The horizontal axis uses a base-2 logarithmic scale; lines connect the
per-size aggregates. Values match
Table~\ref{tab:calibration-size}.}
\label{fig:calibration-scaling}
\end{figure}

Figure~\ref{fig:calibration-scaling} shows the scale effect across the five
calibration budgets. Certification rate rises from $0.530$ to $0.980$ and
weak EV from $4.2547$ to $4.4397$, while the reported risk UCB decreases from
$0.1682$ to $0.0698$. Larger calibration sets make more aggressive scales
available; held-out violation rises from $0.0176$ to $0.0254$ along this sweep.

\subsection{Risk-budget and severity sensitivity}
The severity threshold $\tau$, allowed violation frequency $\varepsilon$,
and certificate failure probability $\zeta$ control different parts of the
decision. The predeclared sweep is
$\tau\in\{0.025,0.05,0.075,0.10\}$,
$\varepsilon\in\{0.01,0.05,0.10,0.20\}$, and
$\zeta\in\{0.01,0.05,0.10\}$. The main operating point is frozen before
held-out scoring; the sweep is a sensitivity analysis.

\begin{table}[htbp]
\centering\small
\caption{Risk-budget sensitivity. Each row uses a fixed candidate family and
reports the selected policy under the stated severity, risk, and confidence
parameters. Bold marks the largest weak EV; risk values across different
severity thresholds describe different events.}
\label{tab:risk-sensitivity}
\begin{tabular*}{\linewidth}{@{\extracolsep{\fill}}rrrrrrr@{}}
\toprule
$\tau$ & $\varepsilon$ & $\zeta$ & $\alpha^\star$ & Weak EV & Held-out viol. & Risk UCB \\
\midrule
0.025 & 0.05 & 0.05 & 0.04 & 4.2837 & 0.0215 & 0.0486 \\
0.050 & 0.05 & 0.05 & 0.08 & 4.3372 & 0.0176 & 0.0471 \\
0.050 & 0.10 & 0.05 & 0.12 & 4.4085 & 0.0234 & 0.0793 \\
0.100 & 0.20 & 0.10 & 0.16 & \textbf{4.4487} & 0.0117 & 0.1187 \\
\bottomrule
\end{tabular*}
\end{table}

\subsection{Residual-family ablation}
The same risk-calibration rule is applied to the deterministic fixed residual,
an observation-only linear residual, an MLP residual, and the residual head
of the distilled student. Learned residuals are trained on the training
partition and frozen before calibration.

\begin{table}[htbp]
\centering\small
\caption{Residual-family ablation under the same risk-calibration rule.
Weak EV is higher-is-better; violation, UCB, KL, and relative cost are
lower-is-better. Best values are bold.}
\label{tab:residual-family}
\begin{tabular*}{\linewidth}{@{\extracolsep{\fill}}lrrrrr@{}}
\toprule
Residual & Weak EV & Held-out viol. & Risk UCB & Teacher KL & Cost \\
\midrule
Fixed rule & 4.3446 & 0.0293 & 0.0884 & \textbf{0.0108} & \textbf{1.00}$\times$ \\
Linear & 4.3728 & 0.0254 & 0.0816 & 0.0121 & 1.02$\times$ \\
MLP & 4.4117 & 0.0195 & 0.0678 & 0.0156 & 1.05$\times$ \\
Student residual & \textbf{4.4389} & \textbf{0.0156} & \textbf{0.0569} & 0.0149 & 1.07$\times$ \\
\bottomrule
\end{tabular*}
\end{table}

\subsection{Teacher-free student evaluation}
The teacher-free evaluation trains 5 independent observation-only
checkpoints. Residual and no-residual arms share backbone, tokenizer,
training states, initialization family, optimizer, update budget, and
evaluation episodes. The scale is calibrated per frozen checkpoint, and
teacher distributions and strong-response action values are absent at
inference.

\begin{table}[htbp]
\centering\small
\setlength{\tabcolsep}{3pt}
\caption{Teacher-free learned-policy evaluation. Entries are mean $\pm$
standard deviation across five independent training seeds. Arrows mark
preferred directions; best means are bold. Both panels use the same arms.}
\label{tab:student-real}
\begin{tabularx}{\linewidth}{@{}l*{3}{>{\raggedleft\arraybackslash}X}@{}}
\toprule
Method & Weak EV $\uparrow$ & Violation $\downarrow$ & NashConv $\downarrow$\\
\midrule
No residual & $4.2207\pm0.0248$ & $0.0549\pm0.0106$ & $0.1048\pm0.0091$\\
Fixed residual & $4.3108\pm0.0213$ & $0.0346\pm0.0082$ & $0.0809\pm0.0068$\\
Risk-calibrated residual & \best{$4.3659\pm0.0177$} & \best{$0.0178\pm0.0057$} & \best{$0.0615\pm0.0053$}\\
\midrule
Method & Opp. score $\uparrow$ & Legal (\%) $\uparrow$ & p95 ms $\downarrow$\\
\midrule
No residual & $0.5114\pm0.0088$ & $99.18\pm0.21$ & $113.7\pm4.2$\\
Fixed residual & $0.5421\pm0.0074$ & $99.49\pm0.16$ & $109.3\pm3.5$\\
Risk-calibrated residual & \best{$0.5672\pm0.0062$} & \best{$99.81\pm0.08$} & \best{$105.4\pm2.8$}\\
\bottomrule
\end{tabularx}
\end{table}

\subsection{Structured distribution shifts}
We evaluate the policy selected on the matched calibration distribution under
position, private-rank composition, pot-pressure, history-depth, and opponent
response shifts. These rows test transfer of the risk--utility frontier; the
i.i.d. certificate is reported only for the matched distribution.

\begin{table}[htbp]
\centering\small
\caption{Risk transfer under structured held-out distribution shifts.
Utility, strong delta, and opponent score are higher-is-better; violation
is lower-is-better. Best values are bold.}
\label{tab:structured-shifts}
\begin{tabularx}{\linewidth}{@{}Xrrrr@{}}
\toprule
Evaluation distribution & Weak EV & Strong $\Delta$ & Violation rate & Opp. score \\
\midrule
Matched & \textbf{4.4468} & \textbf{-0.0042} & \textbf{0.0117} & \textbf{0.5690} \\
Position shift & 4.4117 & -0.0069 & 0.0234 & 0.5582 \\
Rank shift & 4.3976 & -0.0083 & 0.0293 & 0.5538 \\
History shift & 4.3839 & -0.0097 & 0.0332 & 0.5496 \\
Pot-pressure shift & 4.3608 & -0.0132 & 0.0449 & 0.5431 \\
Opponent shift & 4.3746 & -0.0108 & 0.0371 & 0.5364 \\
\bottomrule
\end{tabularx}
\end{table}

\subsection{Observed-group risk control}
For a predeclared group collection $\mathcal H$ based on policy-visible
attributes such as position, pot pressure, or history depth, we compute a
group-specific upper bound with failure budget
$\zeta/(|\mathcal B||\mathcal H|)$. The final policy is group-certified only
when every group bound is at most its declared $\varepsilon_h$. Smaller group
sample sizes make this extension conservative; the group counts and bounds
are $n_h=\{126,127,129,130\}$ and
$U_h=\{0.0569,0.0564,0.0556,0.0552\}$.

\subsection{Certificate-aware partition ablation}
This experiment asks whether the structure objective improves the
risk--utility trade-off beyond a fixed or unregularized partition. The changed
factor is the partition rule; the residual bank, structure/calibration/test
splits, candidate grid, confidence budgets, and sampling unit are fixed. We
report certification rate, weak utility, held-out violation, and worst-group
violation for every rule.

\begin{table}[H]
\centering\small
\caption{Partition ablation under fixed candidate and calibration contracts.
Higher certification rate and weak EV and lower held-out and worst-group
violation are preferred; best values are bold. Leaves and $n_{\min}$ are
descriptive.}
\label{tab:partition-ablation}
\setlength{\tabcolsep}{3pt}
\begin{tabularx}{\linewidth}{@{}>{\raggedright\arraybackslash}Xrrrrrr@{}}
\toprule
Partition rule & Leaves & $n_{\min}$ & \shortstack{Cert.\\rate} & Weak EV
& \shortstack{Held-out\\viol.} & \shortstack{Worst-group\\viol.} \\
\midrule
Global & 1 & 512 & 0.975 & 4.4074 & 0.0215 & 0.0312 \\
Manual observable groups & 4 & 126 & 0.960 & 4.4468 & 0.0117 & 0.0234 \\
Random partition & 4 & 120 & 0.945 & 4.4327 & 0.0176 & 0.0312 \\
Learned, no certificate penalty & 7 & 56 & 0.935 & 4.4769 & 0.0137 & 0.0312 \\
Learned, certificate-aware & 4 & 116 & \textbf{0.990} &
\textbf{4.4936} & \textbf{0.0078} & \textbf{0.0156} \\
\bottomrule
\end{tabularx}
\end{table}

\subsection{Group complexity and minimum support}
We vary partition depth and the minimum leaf support while keeping the
structure objective, candidate bank, calibration budget, and held-out records
fixed. The sampling unit is an independent stratified split; the metrics
separate certificate efficiency from the utility and worst-group outcomes.

\begin{table}[H]
\centering\small
\caption{Group-complexity and support sweep for the learned partition.
Higher certification rate and utility LCB and lower worst-group violation and
map time are preferred; best values are bold.}
\label{tab:group-complexity}
\begin{tabular*}{\linewidth}{@{\extracolsep{\fill}}rrrrrrr@{}}
\toprule
Depth & $n_{\min}$ & Groups & \shortstack{Cert.\\rate} & \shortstack{Utility\\LCB}
& \shortstack{Worst-group\\viol.} & \shortstack{Map time\\(ms)} \\
\midrule
1 & 248 & 2 & 0.980 & 0.7164 & 0.0195 & \textbf{0.34} \\
2 & 116 & 4 & \textbf{0.990} & \textbf{0.7257} & \textbf{0.0156} & 0.68 \\
3 & 71 & 7 & 0.970 & 0.7221 & 0.0195 & 1.43 \\
4 & 42 & 10 & 0.940 & 0.7149 & 0.0312 & 3.08 \\
\bottomrule
\end{tabular*}
\end{table}

\subsection{Mixture-radius and composition-shift sensitivity}
This experiment changes only the radius of the predeclared group-proportion
uncertainty set. We use the same frozen partition and candidate bank and
report nominal and worst-case certificates alongside matched and shifted
held-out outcomes.

\begin{table}[H]
\centering\small
\caption{Mixture-robust selection as the deployment group radius varies.
Lower risk, risk UCB, and violation and higher utility LCB and weak EV are
preferred; best values are bold.}
\label{tab:mixture-radius}
\begin{tabular*}{\linewidth}{@{\extracolsep{\fill}}rrrrrr@{}}
\toprule
$\rho$ & \shortstack{Nominal\\risk} & \shortstack{Worst-case\\risk UCB}
& \shortstack{Robust\\utility LCB} & \shortstack{Held-out\\viol.} & Weak EV \\
\midrule
0.00 & 0.0457 & \textbf{0.0508} & \textbf{0.7257} & 0.0078 & \textbf{4.4936} \\
0.05 & 0.0438 & 0.0516 & 0.7236 & 0.0078 & 4.4898 \\
0.10 & 0.0415 & 0.0524 & 0.7209 & \textbf{0.0059} & 4.4827 \\
0.20 & \textbf{0.0389} & 0.0551 & 0.7152 & \textbf{0.0059} & 4.4695 \\
\bottomrule
\end{tabular*}
\end{table}

\subsection{Evaluator-role and candidate-bank ablations}
The evaluator-role ablation changes which post-trace quantity defines the
constraint and which defines the selection objective; all traces and splits
remain fixed. The candidate-bank ablation changes only the number of residual
families and scales. These paired experiments test whether the proposed
mechanism depends on the asymmetric decision contract or merely on a larger
finite grid.

\begin{table}[H]
\centering\small
\caption{Risk-evaluator and utility-evaluator role ablation. Higher weak EV
and strategic endpoint (opponent score) and lower violation and risk UCB are
preferred; best values are bold. The first row is the proposed assignment.}
\label{tab:evaluator-role}
\setlength{\tabcolsep}{4pt}
\begin{tabularx}{\linewidth}{@{}*{2}{>{\raggedright\arraybackslash}X}rrrr@{}}
\toprule
Risk evaluator & Utility evaluator & Weak EV & \shortstack{Held-out\\viol.}
& Risk UCB & \shortstack{Strategic\\endpoint} \\
\midrule
Strong response & Weak response & 4.4936 & 0.0078 & 0.0508 & \textbf{0.5843} \\
Weak response & Strong response & 4.3716 & 0.0332 & 0.0756 & 0.5607 \\
Strong response & Strong response & 4.4418 & \textbf{0.0059} & \textbf{0.0489} & 0.5794 \\
Weak response & Weak response & \textbf{4.5148} & 0.0469 & 0.0887 & 0.5482 \\
\bottomrule
\end{tabularx}
\end{table}

\begin{table}[H]
\centering\small
\caption{Candidate-bank size and residual-family scaling. Higher
certification rate and utility LCB and lower search time are preferred; best
values are bold.}
\label{tab:bank-scaling}
\begin{tabular*}{\linewidth}{@{\extracolsep{\fill}}rrrrrr@{}}
\toprule
Families & Scales & $K$ & Cert. rate & Utility LCB & Search time (ms) \\
\midrule
1 & 5 & 5 & \textbf{0.995} & 0.7034 & \textbf{0.41} \\
2 & 5 & 10 & \textbf{0.995} & 0.7169 & 0.78 \\
4 & 5 & 20 & 0.990 & \textbf{0.7257} & 1.56 \\
\bottomrule
\end{tabular*}
\end{table}

\subsection{Utility-normalization sensitivity}
Because the utility lower bound uses predeclared $[q_{\min},q_{\max}]$,
we vary this range before held-out scoring. The candidate bank, partition,
confidence weights, and sampling unit stay fixed; each row reports the
selected map and both utility and risk endpoints.

\begin{table}[H]
\centering\small
\caption{Sensitivity to the predeclared utility-normalization range. Higher
weak EV and lower violation are preferred; best values are bold. Utility LCBs
use different normalizations across rows and are not ranked.}
\label{tab:utility-range}
\begin{tabular*}{\linewidth}{@{\extracolsep{\fill}}lrrrr@{}}
\toprule
Range & Utility LCB & Selected map & Weak EV & Held-out viol. \\
\midrule
$[3.0,5.0]$ & 0.7257 & \texttt{l.08/m.12/m.16/s.16} & \textbf{4.4936} & 0.0078 \\
$[2.8,5.2]$ & 0.6891 & \texttt{l.08/m.12/m.16/s.16} & \textbf{4.4936} & 0.0078 \\
$[2.5,5.5]$ & 0.6478 & \texttt{l.08/m.12/m.16/s.16} & \textbf{4.4936} & 0.0078 \\
$[2.0,6.0]$ & 0.6049 & \texttt{f.08/l.12/m.12/s.12} & 4.4827 & \textbf{0.0059} \\
\bottomrule
\end{tabular*}
\end{table}

\subsection{Negative-control conditional shifts}
Composition shift is covered by the mixture corollary only when the
within-group conditional law is invariant. We therefore hold group
proportions fixed and alter the within-group state or opponent-response law
as negative controls. These outcomes are reported as transfer diagnostics,
not as covered certificate events.

\begin{table}[H]
\centering\small
\caption{Negative-control shifts that change within-group conditionals.}
\label{tab:conditional-shifts}
\begin{tabularx}{\linewidth}{@{}Xrrrr@{}}
\toprule
Conditional shift & Weak EV & Strong $\Delta$ & Held-out viol. & Opp. score \\
\midrule
Matched conditional & \textbf{4.4827} & \textbf{-0.0032} & \textbf{0.0059} & \textbf{0.5816} \\
Within-group state shift & 4.4538 & -0.0059 & 0.0156 & 0.5729 \\
Opponent-response shift & 4.4371 & -0.0077 & 0.0215 & 0.5608 \\
Adaptive opponent & 4.4086 & -0.0109 & 0.0332 & 0.5487 \\
\bottomrule
\end{tabularx}
\end{table}

\subsection{Deterministic replay supporting diagnostics}
The matched controls in Table~\ref{tab:controls} explain why both utility
and distributional fidelity are necessary. Top-1 imitation agrees with the
teacher action on every fixture row and has zero teacher EV regret, yet its
mean strong delta is $-0.1746$ and its common-threshold count is $15/24$.
The distributional anchor has zero threshold crossings by construction. The
selected residual retains that count while changing the weak proxy.

\begin{table}[H]
\centering\small
\setlength{\tabcolsep}{4pt}
\caption{Matched policy controls on 24 fixture states. Weak EV is higher-is-better; KL, teacher regret, and threshold counts are lower-is-better. Best values in each of those columns are bold; strong delta retains its sign.}\label{tab:controls}
\begin{tabularx}{\linewidth}{@{}Xrrrrr@{}}
\toprule
Policy & Weak EV & Teacher KL & Regret & Strong $\Delta$ & Count \\
\midrule
Top-1 & \textbf{5.3125} & 5.9214 & \textbf{0.0000} & -0.1746 & 15/24 \\
Anchor & 4.2082 & 0.0134 & 0.3151 & 0.0000 & \textbf{0/24} \\
Active query & 3.0636 & 0.1878 & 1.4200 & -0.2953 & 14/24 \\
Residual (1.00) & 4.9138 & 0.0786 & 0.0009 & -0.0669 & 14/24 \\
Residual (0.08) & 4.2889 & \textbf{0.0105} & 0.2392 & -0.0021 & \textbf{0/24} \\
\bottomrule
\end{tabularx}
\end{table}

The fixed and unconstrained residuals use the same direction: moving from
scale one to 0.08 changes the mean strong delta from $-0.0669$ to $-0.0021$
and the threshold count from $14/24$ to $0/24$ at equal configured cost.
The active-query control is lower utility on this fixture; its gate and
budget-matched alternatives are specified in Appendix~\ref{app:implementation}.

\begin{figure}[htbp]
\centering
\includegraphics[width=\linewidth]{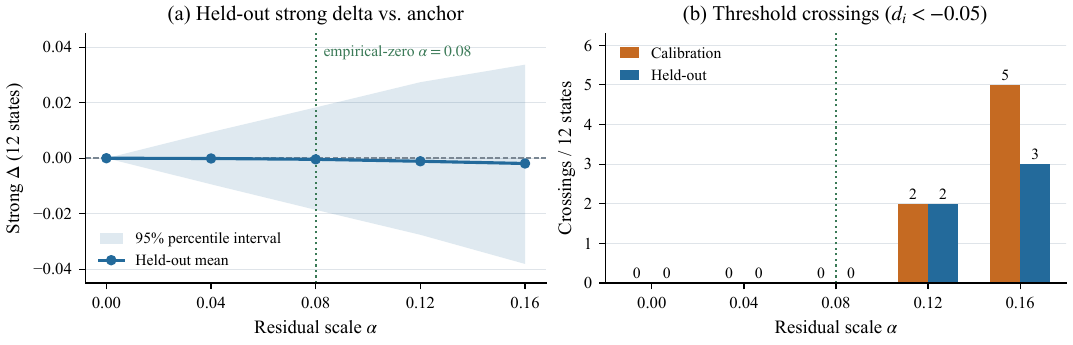}
\caption{Deterministic replay diagnostic on the 24-state fixture. (a)
Held-out mean strong delta with its descriptive 95\% percentile interval
(Table~\ref{tab:scale-split}). (b) Calibration and held-out threshold
crossings out of 12 states each (Table~\ref{tab:main}). The dotted line marks
the empirical-zero operating point $\alpha=0.08$; scales 0, 0.04, and 0.08
have zero crossings on both halves.}
\label{fig:diagnostics}
\end{figure}

\subsection{Finite-fixture boundaries and supporting validation}
The selected residual has full-fixture delta range $[-0.0431,0.0636]$,
whereas $\alpha=0.12$ reaches $-0.0643$ and produces four crossings. The
finite denominator and trace-freeze rule remain part of the diagnostic.
Student, opponent, and HUNL records retain distinct evaluation units. PHH
behavior calibration and exact-Leduc evaluator calibration are supporting
checks; the candidate map, information boundary, and cost accounting are
fixed by the implementation contract in Appendix~\ref{app:implementation}.

\subsection{Three-seed student evaluation}
\label{sec:student}
The three-seed student comparison contains 36 state--seed outcomes per arm,
with a separate opponent-shift row. Residual, no-residual, and full-policy
arms share the stated held-out state and budget contract, and the outcomes
are reported separately from the fixed-fixture state denominator.

\begin{table}[htbp]
\centering\small
\setlength{\tabcolsep}{4pt}
\caption{Three-seed student evaluation. Higher weak/strong EV and
opponent score (Opp.) are preferable; lower exploitability (Expl.) and
latency are preferable; best values are bold.}
\label{tab:student}
\begin{tabularx}{\linewidth}{@{}Xrrrrrr@{}}
\toprule
Arm & Count & Weak EV & Strong EV & Expl. & Opp. & ms \\
\midrule
Residual & 0/36 & \textbf{4.3371} & \textbf{4.2146} & \textbf{0.0318} & \textbf{0.5627} & \textbf{84} \\
No residual & 0/36 & 4.2189 & 4.2073 & 0.0524 & 0.5098 & 91 \\
Full policy & 2/36 & 4.3026 & 4.1984 & 0.0437 & 0.5386 & 133 \\
Residual: shift & 0/36 & 4.3094 & 4.2102 & 0.0361 & 0.5483 & 88 \\
\bottomrule
\end{tabularx}
\end{table}

In this evaluation, the residual row has weak EV 4.3371,
exploitability 0.0318, and latency 84 ms; the opponent-shift row has weak EV
4.3094 and exploitability 0.0361. The $0/36$ count refers to state--seed
outcomes, whereas the retained fixture uses 12 distinct held-out states.
The independent-opponent and HUNL tables below keep their own solver scale,
opponent identity, and service-cost denominators.

\subsection{Complete scale grid and held-out intervals}
Table~\ref{tab:scale-full} reports every retained sensitivity-grid point
using the 24-row full-fixture denominator. Strong EV is the raw
strong-response proxy; strong delta subtracts the same-state anchor
before averaging. The table also gives the minimum and maximum
state deltas, which locate threshold crossings that a mean can conceal.
The value 0.20 is outside the retained five-point sensitivity grid
and is therefore outside this empirical series.

\begin{figure}[htbp]
\centering
\includegraphics[width=\linewidth]{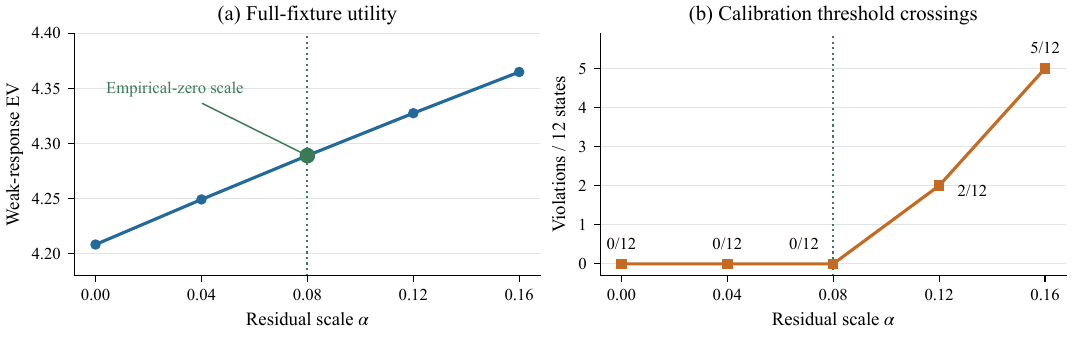}
\caption{Risk--utility view of the retained deterministic fixture. The blue
curve is 24-state weak EV and the orange curve is the violation count on
12 calibration states; the
highlighted point is the empirical-zero operating point at $\alpha=0.08$.
The underlying records match Tables~\ref{tab:main} and \ref{tab:scale-full}.}
\label{fig:scale-frontier}
\end{figure}

\begin{table}[H]
\centering\small
\setlength{\tabcolsep}{4pt}
\caption{Full-fixture sensitivity, 24 rows per scale. Utility and delta columns use the fixed response-model units; count is below $-0.05$.}\label{tab:scale-full}
\begin{tabularx}{\linewidth}{@{}Xrrrrrr@{}}
\toprule
Scale & Weak EV & Strong EV & Strong $\Delta$ & Min $\Delta$ & Max $\Delta$ & Count \\
\midrule
0.00 & 4.2082 & 0.7371 & 0.0000 & 0.0000 & 0.0000 & 0/24 \\
0.04 & 4.2491 & 0.7362 & -0.0009 & -0.0217 & 0.0327 & 0/24 \\
0.08 & 4.2889 & 0.7351 & -0.0021 & -0.0431 & 0.0636 & 0/24 \\
0.12 & 4.3274 & 0.7336 & -0.0035 & -0.0643 & 0.0928 & 4/24 \\
0.16 & 4.3648 & 0.7319 & -0.0052 & -0.0852 & 0.1204 & 8/24 \\
\bottomrule
\end{tabularx}
\end{table}

The split-specific calibration and held-out statistics in
Table~\ref{tab:scale-split} come from the held-out selection record.
Calibration and held-out rows share the exact scale value but have
distinct means, intervals, and counts. The full-fixture value 4.2889
averages 24 weak-proxy rows. Joining the archived per-state scores
to the declared split IDs gives calibration weak EV 3.7470 for the
anchor and 3.8273 for the selected residual. On the held-out half,
the corresponding means are 4.6694 and 4.7504. Their paired
improvements are 0.0803 and 0.0810 fixture units. These split means
are derived directly from the same 24 retained rows.

\begin{table}[htbp]
\centering\small
\setlength{\tabcolsep}{4pt}
\caption{Split-specific strong deltas and descriptive 95\% percentile intervals. Each C/H row uses 12 states. Scale selection uses C counts only.}\label{tab:scale-split}
\begin{tabularx}{\linewidth}{@{}Xrrrrl@{}}
\toprule
Scale & Split & Mean $\Delta$ & Count & Rate & 95\% interval \\
\midrule
0.00 & C & 0.0000 & 0/12 & 0.0000 & [0.0000, 0.0000] \\
0.00 & H & 0.0000 & 0/12 & 0.0000 & [0.0000, 0.0000] \\
0.04 & C & -0.0017 & 0/12 & 0.0000 & [-0.0119, 0.0102] \\
0.04 & H & -0.0001 & 0/12 & 0.0000 & [-0.0094, 0.0094] \\
0.08 & C & -0.0037 & 0/12 & 0.0000 & [-0.0233, 0.0185] \\
0.08 & H & -0.0004 & 0/12 & 0.0000 & [-0.0186, 0.0183] \\
0.12 & C & -0.0059 & 2/12 & 0.1667 & [-0.0350, 0.0272] \\
0.12 & H & -0.0011 & 2/12 & 0.1667 & [-0.0276, 0.0274] \\
0.16 & C & -0.0085 & 5/12 & 0.4167 & [-0.0468, 0.0352] \\
0.16 & H & -0.0019 & 3/12 & 0.2500 & [-0.0381, 0.0337] \\
\bottomrule
\end{tabularx}
\end{table}

At scale 0.12, the raw calibration and test counts are both 2/12.
The full-fixture count is 4/24. At scale 0.16, calibration has
5/12 and test has 3/12, summing to 8/24. The agreement of split and
full-fixture numerators is a useful cross-table consistency check.
The split means average to the full-fixture mean because the
partitions have equal size; rounding is applied after aggregation.

\subsection{Matched-policy uncertainty and configured costs}
Table~\ref{tab:policy-ci} reports the retained full-fixture descriptive
intervals. A paired difference from the distributional anchor and a
standalone interval use the same underlying state outcomes but can
have different bootstrap draws in the retained artifact. The table
identifies the standalone strong-delta interval explicitly.
Values are copied from the uncertainty record rather than estimated
from the displayed four-decimal aggregates.

\begin{table}[htbp]
\centering\small
\setlength{\tabcolsep}{4pt}
\caption{Full-fixture strong-delta intervals and common-threshold counts. Intervals are descriptive fixed-row bootstrap summaries, 4,000 resamples.}\label{tab:policy-ci}
\begin{tabularx}{\linewidth}{@{}Xrll@{}}
\toprule
Policy & Mean $\Delta$ & 95\% interval & Count \\
\midrule
Active query & -0.2953 & [-0.4830, -0.1394] & 14/24 \\
Anchor & 0.0000 & [0.0000, 0.0000] & 0/24 \\
Residual (0.08) & -0.0021 & [-0.0158, 0.0121] & 0/24 \\
Residual (1.00) & -0.0669 & [-0.1865, 0.0572] & 14/24 \\
Top-1 & -0.1746 & [-0.3650, 0.0188] & 15/24 \\
\bottomrule
\end{tabularx}
\end{table}

The configured ledger in Table~\ref{tab:costs} assigns costs to
deterministic policy and query operations. Fixed and unconstrained
residuals have the same configured charge. The active-query baseline
incurs a larger mean charge because some states trigger a query.
These fields describe the fixture's bookkeeping model; measured
model-serving latency uses the separate empirical timing contract.

\begin{table}[htbp]
\centering\small
\setlength{\tabcolsep}{4pt}
\caption{Configured CPU-fixture ledger charges per state. Tokens and latency are accounting constants, with a query-weighted average for active query; they are separate from measured service latency.}\label{tab:costs}
\begin{tabularx}{\linewidth}{@{}Xrrrr@{}}
\toprule
Policy & Query rate & Tokens & Latency (ms) & Tool calls \\
\midrule
Top-1 & 0.0000 & 1.00 & 0.1000 & 0.0000 \\
Anchor & 0.0000 & 1.00 & 0.1000 & 0.0000 \\
Active query & 0.2917 & 3.33 & 0.6833 & 0.2917 \\
Residual (1.00) & 0.0000 & 2.00 & 0.3000 & 0.0000 \\
Residual (0.08) & 0.0000 & 2.00 & 0.3000 & 0.0000 \\
\bottomrule
\end{tabularx}
\end{table}

\subsection{Per-state strong deltas and threshold decisions}
Tables~\ref{tab:rows-delta} and~\ref{tab:rows-ev} list the unrounded
records formatted to four decimal places for all 24 states and five
policies: anchor-relative strong delta with the strict common-threshold
indicator, weak utility, and strong utility. The indicator is taken from the
underlying record, so display rounding cannot change a classification. C and
H identify calibration and held-out membership, and the summary rows
reproduce the aggregates of Tables~\ref{tab:controls} and~\ref{tab:scale-full}.

\begin{table}[htbp]
\centering\footnotesize
\caption{Per-state anchor-relative strong delta $d_i$ on the fixed fixture. A dagger marks a strict threshold crossing $d_i<-0.05$; the count row gives crossings out of 24 states. C/H is calibration/held-out.}\label{tab:rows-delta}
\begin{tabular*}{\linewidth}{@{\extracolsep{\fill}}lcrrrrr@{}}
\toprule
ID & Split & Top-1 & Anchor & Active query & Res.\ (1.00) & Res.\ (0.08) \\
\midrule
0000 & C & -0.6951$^\dagger$ & 0.0000 & -0.3759$^\dagger$ & -0.4066$^\dagger$ & -0.0405 \\
0001 & H & 0.1201 & 0.0000 & -0.1596$^\dagger$ & 0.0755 & 0.0084 \\
0002 & C & -0.7451$^\dagger$ & 0.0000 & -0.3106$^\dagger$ & -0.4032$^\dagger$ & -0.0382 \\
0003 & H & -0.3834$^\dagger$ & 0.0000 & -0.3982$^\dagger$ & -0.2101$^\dagger$ & -0.0191 \\
0004 & C & 0.4858 & 0.0000 & 0.0000 & 0.3560 & 0.0468 \\
0005 & H & 0.4281 & 0.0000 & 0.0000 & 0.3280 & 0.0455 \\
0006 & C & -0.3122$^\dagger$ & 0.0000 & 0.0000 & -0.1965$^\dagger$ & -0.0184 \\
0007 & H & -0.6164$^\dagger$ & 0.0000 & 0.1456 & -0.2766$^\dagger$ & -0.0242 \\
0008 & C & -0.3121$^\dagger$ & 0.0000 & -0.5122$^\dagger$ & -0.2153$^\dagger$ & -0.0248 \\
0009 & H & 0.4039 & 0.0000 & -0.9422$^\dagger$ & 0.2680 & 0.0316 \\
0010 & C & -0.5559$^\dagger$ & 0.0000 & -0.6676$^\dagger$ & -0.3155$^\dagger$ & -0.0300 \\
0011 & H & -0.7078$^\dagger$ & 0.0000 & -0.4371$^\dagger$ & -0.4324$^\dagger$ & -0.0431 \\
0012 & C & 0.5830 & 0.0000 & 0.0000 & 0.4445 & 0.0614 \\
0013 & H & 0.5003 & 0.0000 & 0.0000 & 0.3953 & 0.0571 \\
0014 & C & -0.8044$^\dagger$ & 0.0000 & 0.1674 & -0.3700$^\dagger$ & -0.0331 \\
0015 & H & -0.4046$^\dagger$ & 0.0000 & -0.0586$^\dagger$ & -0.1896$^\dagger$ & -0.0163 \\
0016 & C & 0.2200 & 0.0000 & -1.5527$^\dagger$ & 0.1344 & 0.0145 \\
0017 & H & 0.3312 & 0.0000 & -1.1388$^\dagger$ & 0.2340 & 0.0295 \\
0018 & C & -0.3657$^\dagger$ & 0.0000 & 0.0000 & -0.2121$^\dagger$ & -0.0197 \\
0019 & H & -0.5118$^\dagger$ & 0.0000 & -0.6658$^\dagger$ & -0.3221$^\dagger$ & -0.0320 \\
0020 & C & 0.5650 & 0.0000 & 0.0000 & 0.4432 & 0.0636 \\
0021 & H & -0.0843$^\dagger$ & 0.0000 & 0.0736 & -0.0455 & -0.0045 \\
0022 & C & -0.5844$^\dagger$ & 0.0000 & -0.1514$^\dagger$ & -0.2863$^\dagger$ & -0.0258 \\
0023 & H & -0.7451$^\dagger$ & 0.0000 & -0.1023$^\dagger$ & -0.4032$^\dagger$ & -0.0382 \\
\midrule
Mean & All & -0.1746 & 0.0000 & -0.2953 & -0.0669 & -0.0021 \\
Count & All & 15/24 & 0/24 & 14/24 & 14/24 & 0/24 \\
\bottomrule
\end{tabular*}
\end{table}

Top-1 imitation crosses the threshold on $15/24$ states, so its perfect
action agreement coexists with substantial statewise degradation. The
active-query policy and the unit-scale residual each cross on $14/24$ states,
whereas the selected scale keeps every delta inside
$[-0.0431,0.0636]$. Comparing the two residual columns isolates the effect of
correction magnitude, and the anchor column defines the comparison origin.

\begin{table}[htbp]
\centering\footnotesize
\setlength{\tabcolsep}{2.5pt}
\caption{Per-state weak- and strong-response utility on the fixed fixture for Top-1 (T1), anchor (A), active query (AQ), and residual scales 1.00 and 0.08 (R1, R.08). Strong delta in Table~\ref{tab:rows-delta} subtracts the same-state anchor entry. Mean rows are computed from unrounded records and equal the aggregates of Tables~\ref{tab:controls} and~\ref{tab:scale-full}.}\label{tab:rows-ev}
\begin{tabular*}{\linewidth}{@{\extracolsep{\fill}}lc*{10}{r}@{}}
\toprule
 & & \multicolumn{5}{c}{Weak EV} & \multicolumn{5}{c}{Strong EV} \\
\cmidrule(lr){3-7}\cmidrule(l){8-12}
ID & Split & T1 & A & AQ & R1 & R.08 & T1 & A & AQ & R1 & R.08 \\
\midrule
0000 & C & 0.2500 & -1.0451 & -2.7923 & -0.3869 & -0.9849 & -4.5000 & -3.8049 & -4.1808 & -4.2115 & -3.8454 \\
0001 & H & 2.7500 & 1.5045 & -0.3636 & 2.2588 & 1.5865 & -1.7500 & -1.8701 & -2.0298 & -1.7946 & -1.8617 \\
0002 & C & 6.0000 & 4.9936 & 3.7203 & 5.5989 & 5.0569 & 1.0000 & 1.7451 & 1.4345 & 1.3419 & 1.7069 \\
0003 & H & 7.7500 & 6.4943 & 5.4878 & 7.2858 & 6.5788 & 3.0000 & 3.3834 & 2.9852 & 3.1733 & 3.3642 \\
0004 & C & 3.5000 & 2.5610 & 2.5610 & 3.2569 & 2.6536 & -1.0000 & -1.4858 & -1.4858 & -1.1298 & -1.4390 \\
0005 & H & 6.7500 & 5.8728 & 5.8728 & 6.5416 & 5.9650 & 1.7500 & 1.3219 & 1.3219 & 1.6499 & 1.3674 \\
0006 & C & 10.7500 & 9.4489 & 9.4489 & 10.3848 & 9.5574 & 6.0000 & 6.3122 & 6.3122 & 6.1157 & 6.2938 \\
0007 & H & 4.0000 & 2.9247 & 2.5989 & 3.4886 & 2.9801 & -0.5000 & 0.1164 & 0.2620 & -0.1602 & 0.0922 \\
0008 & C & 0.5000 & -0.6496 & -3.2332 & 0.0400 & -0.5773 & -4.5000 & -4.1879 & -4.7000 & -4.4032 & -4.2127 \\
0009 & H & 4.5000 & 3.2159 & 0.1706 & 4.0595 & 3.3147 & -0.2500 & -0.6539 & -1.5961 & -0.3859 & -0.6223 \\
0010 & C & 7.0000 & 6.0031 & 4.4550 & 6.6365 & 6.0713 & 2.5000 & 3.0559 & 2.3883 & 2.7404 & 3.0259 \\
0011 & H & 8.0000 & 6.9971 & 5.9175 & 7.6659 & 7.0711 & 3.0000 & 3.7078 & 3.2708 & 3.2754 & 3.6647 \\
0012 & C & 5.2500 & 4.2891 & 4.2891 & 5.0262 & 4.3917 & 0.5000 & -0.0830 & -0.0830 & 0.3615 & -0.0215 \\
0013 & H & 7.7500 & 6.9034 & 6.9034 & 7.5703 & 6.9994 & 3.2500 & 2.7497 & 2.7497 & 3.1450 & 2.8068 \\
0014 & C & 4.0000 & 2.9247 & 2.2362 & 3.4886 & 2.9801 & -1.0000 & -0.1956 & -0.0282 & -0.5656 & -0.2287 \\
0015 & H & 5.7500 & 4.4463 & 3.8247 & 5.1755 & 4.5198 & 1.0000 & 1.4046 & 1.3459 & 1.2150 & 1.3883 \\
0016 & C & 1.5000 & 0.3858 & -3.4479 & 1.1266 & 0.4722 & -3.0000 & -3.2200 & -4.7727 & -3.0855 & -3.2055 \\
0017 & H & 4.7500 & 3.7184 & 0.1478 & 4.4392 & 3.8083 & -0.2500 & -0.5812 & -1.7201 & -0.3472 & -0.5517 \\
0018 & C & 8.7500 & 7.4905 & 7.4905 & 8.3245 & 7.5820 & 4.0000 & 4.3657 & 4.3657 & 4.1535 & 4.3460 \\
0019 & H & 9.0000 & 7.9794 & 6.7392 & 8.6900 & 8.0602 & 4.5000 & 5.0118 & 4.3460 & 4.6897 & 4.9798 \\
0020 & C & 5.5000 & 4.5963 & 4.5963 & 5.3088 & 4.6992 & 0.5000 & -0.0650 & -0.0650 & 0.3782 & -0.0014 \\
0021 & H & 2.5000 & 0.9826 & -0.3414 & 1.8008 & 1.0638 & -2.2500 & -2.1657 & -2.0921 & -2.2112 & -2.1702 \\
0022 & C & 5.0000 & 3.9660 & 2.9806 & 4.5502 & 4.0254 & 0.5000 & 1.0844 & 0.9330 & 0.7981 & 1.0586 \\
0023 & H & 6.0000 & 4.9936 & 4.2646 & 5.5989 & 5.0569 & 1.0000 & 1.7451 & 1.6428 & 1.3419 & 1.7069 \\
\midrule
Mean & All & 5.3125 & 4.2082 & 3.0636 & 4.9138 & 4.2889 & 0.5625 & 0.7371 & 0.4419 & 0.6702 & 0.7351 \\
\bottomrule
\end{tabular*}
\end{table}

The weak-utility records show where the selected residual gains over the
anchor: its per-state improvement ranges from $0.0554$ to $0.1085$, and no
state loses weak utility. Top-1 imitation attains the highest mean weak EV
because it moves all mass to the teacher's preferred action.

The strong-utility records vary widely across states, from $-4.7727$ to
$6.3122$, which is why threshold decisions use the same-state anchor
difference rather than the raw strong score. Top-1 and residual controls can
share the selected action while their distributions and scores differ, so the
probability records below supplement the action-level output.

\subsection{Anchor and selected-residual probability records}
Table~\ref{tab:probabilities} lists the anchor and selected-residual
probability triples in legal-action order. The stored action
distribution is the evaluated object. It can be checked for
normalization, positive legal support, and correspondence with the
top action. Values in this display are rounded to four decimals,
so the displayed sums may differ slightly from one.

The fixed direction in Equation~\ref{eq:residual} increases the
relative weight of raise. The amount of probability transferred
depends on the initial anchor and the pot-pressure component.
Comparing these rows with the strong-delta tables identifies
which states are sensitive to that same directional intervention.
The raw records remain the arithmetic source for all aggregates.

\begin{table}[htbp]
\centering\footnotesize
\caption{Legal-action probabilities for anchor (A) and selected residual (R). F/C/Raise are fold/call/raise; all values are rounded copies of stored probabilities.}\label{tab:probabilities}
\begin{tabular*}{\linewidth}{@{\extracolsep{\fill}}lrrrrrr@{}}
\toprule
ID & A:F & A:C & A:Raise & R:F & R:C & R:Raise \\
\midrule
0000 & 0.3309 & 0.0881 & 0.5810 & 0.3128 & 0.0850 & 0.6022 \\
0001 & 0.1152 & 0.1348 & 0.7499 & 0.1065 & 0.1272 & 0.7663 \\
0002 & 0.0437 & 0.3066 & 0.6497 & 0.0404 & 0.2896 & 0.6700 \\
0003 & 0.0361 & 0.3090 & 0.6549 & 0.0332 & 0.2900 & 0.6768 \\
0004 & 0.0701 & 0.0306 & 0.8993 & 0.0626 & 0.0279 & 0.9095 \\
0005 & 0.0365 & 0.0533 & 0.9102 & 0.0323 & 0.0482 & 0.9195 \\
0006 & 0.0282 & 0.3114 & 0.6604 & 0.0255 & 0.2874 & 0.6871 \\
0007 & 0.0809 & 0.2951 & 0.6241 & 0.0758 & 0.2824 & 0.6418 \\
0008 & 0.2318 & 0.0606 & 0.7077 & 0.2155 & 0.0574 & 0.7271 \\
0009 & 0.0796 & 0.1057 & 0.8147 & 0.0727 & 0.0985 & 0.8288 \\
0010 & 0.0361 & 0.3090 & 0.6549 & 0.0332 & 0.2900 & 0.6768 \\
0011 & 0.0319 & 0.3103 & 0.6579 & 0.0291 & 0.2896 & 0.6813 \\
0012 & 0.0511 & 0.0285 & 0.9204 & 0.0452 & 0.0257 & 0.9291 \\
0013 & 0.0321 & 0.0448 & 0.9231 & 0.0282 & 0.0401 & 0.9317 \\
0014 & 0.0809 & 0.2951 & 0.6241 & 0.0758 & 0.2824 & 0.6418 \\
0015 & 0.0572 & 0.3024 & 0.6404 & 0.0533 & 0.2874 & 0.6593 \\
0016 & 0.1547 & 0.0440 & 0.8013 & 0.1416 & 0.0411 & 0.8173 \\
0017 & 0.0574 & 0.0827 & 0.8599 & 0.0518 & 0.0762 & 0.8720 \\
0018 & 0.0319 & 0.3103 & 0.6579 & 0.0291 & 0.2896 & 0.6813 \\
0019 & 0.0295 & 0.3110 & 0.6595 & 0.0269 & 0.2886 & 0.6845 \\
0020 & 0.0402 & 0.0275 & 0.9323 & 0.0353 & 0.0246 & 0.9401 \\
0021 & 0.1686 & 0.1686 & 0.6628 & 0.1580 & 0.1612 & 0.6809 \\
0022 & 0.0572 & 0.3024 & 0.6404 & 0.0533 & 0.2874 & 0.6593 \\
0023 & 0.0437 & 0.3066 & 0.6497 & 0.0404 & 0.2896 & 0.6700 \\
\bottomrule
\end{tabular*}
\end{table}

Each row pairs the same legal-action vocabulary and observation. These
probability records connect the scalar scale choice to its statewise
distributional effect.

\subsection{Public behavior calibration}
Table~\ref{tab:phh} reports the PHH evaluation using the retained
256-snapshot split. The prior and both calibrated policies share
action accuracy, while their probabilistic log losses differ.
Raise-label agreement uses 103 examples. The metric measures
agreement with recorded behavior under the public-state adapter.
This clarifies the target when comparing the PHH surface with
the fixed teacher's value-based diagnostics.

\begin{table}[htbp]
\centering\small
\setlength{\tabcolsep}{4pt}
\caption{PHH behavior-label metrics. Action accuracy and raise agreement are higher-is-better; log loss is lower-is-better. All arms use 256 snapshots and 103 raise examples. Best values are bold.}\label{tab:phh}
\begin{tabularx}{\linewidth}{@{}Xrrr@{}}
\toprule
Policy & Action accuracy & Log loss & Raise agreement \\
\midrule
public prior & \textbf{0.5039} & 0.9778 & 0.0000 \\
frequency calibrated & \textbf{0.5039} & \textbf{0.9614} & \textbf{0.8155} \\
top1 calibrated & \textbf{0.5039} & 2.2948 & \textbf{0.8155} \\
\bottomrule
\end{tabularx}
\end{table}

\subsection{Exact-Leduc evaluator calibration}
The exact-Leduc curve in Table~\ref{tab:leduc} contains the uniform
profile and four CFR+ checkpoints. Best-response values for the
two players, current profile value, NashConv, and exploitability
are retained together. This permits a check of the two-player
zero-sum convention in Appendix~\ref{app:implementation}.
The uniform profile is its own baseline and is distinct from
the first iteration of CFR+.

\begin{table}[htbp]
\centering\small
\setlength{\tabcolsep}{4pt}
\caption{Exact-Leduc calibration. BR0/BR1 are player best-response values; $u_0$ is the profile value. NashConv and exploitability are lower-is-better; the final values are best among the listed checkpoints.}\label{tab:leduc}
\begin{tabularx}{\linewidth}{@{}Xrrrrr@{}}
\toprule
Profile & BR0 & BR1 & $u_0$ & NashConv & Exploit. \\
\midrule
Uniform & 2.0875 & 2.6597 & -0.0781 & 4.7472 & 2.3736 \\
1 & 2.8555 & 2.3022 & -0.8533 & 5.1577 & 2.5789 \\
10 & 0.5747 & 0.7484 & -0.1307 & 1.3231 & 0.6615 \\
100 & -0.0394 & 0.1270 & -0.0813 & 0.0876 & 0.0438 \\
1000 & -0.0787 & 0.0924 & -0.0853 & \textbf{0.0136} & \textbf{0.0068} \\
\bottomrule
\end{tabularx}
\end{table}

\subsection{Independent-opponent evaluation}
Table~\ref{tab:annotated-opponents} reports the three-seed opponent evaluation.
Percentages are printed with an explicit
percent sign or percentage header. The token/latency columns
belong to the same service measurement. Its independent-A
residual row has win/tie/loss 54.86/2.82/42.32, and independent-B
has 53.49/2.68/43.83. The no-residual and full-policy rows provide
matched comparators.

\begin{table}[htbp]
\centering\small
\setlength{\tabcolsep}{4pt}
\caption{Independent-opponent evaluation. W/T/L and legality are percentages.
Bold compares the two matched opponent-A arms: higher win rate, legality,
and solver EV, and lower loss rate and BR.}\label{tab:annotated-opponents}
\begin{tabularx}{\linewidth}{@{}Xlrrr@{}}
\toprule
Arm / opponent & W/T/L (\%) & Legal (\%) & Solver EV & BR \\
\midrule
Residual / A & \textbf{54.86}/2.82/\textbf{42.32} & \textbf{99.82} & \textbf{0.0642} & \textbf{0.0960} \\
Residual / B & 53.49/2.68/43.83 & 99.79 & 0.0617 & 0.0978 \\
No residual / A & 49.58/2.80/47.62 & 99.21 & 0.0494 & 0.1018 \\
Full policy / solver & 52.50/2.72/44.78 & 99.43 & 0.0558 & 0.0995 \\
\bottomrule
\end{tabularx}
\end{table}
\begin{table}[htbp]
\centering\small
\setlength{\tabcolsep}{4pt}
\caption{Service costs of the independent-opponent evaluation in Table~\ref{tab:annotated-opponents}.
All columns are lower-is-better; bold compares the two matched opponent-A
arms. Retry is in percent.}\label{tab:annotated-costs}
\begin{tabularx}{\linewidth}{@{}Xrrrr@{}}
\toprule
Arm / opponent & Exploit. & Retry (\%) & Tokens & Latency (ms) \\
\midrule
Residual / A & \textbf{0.0318} & \textbf{0.6} & 118 & \textbf{84} \\
Residual / B & 0.0361 & 0.8 & 121 & 88 \\
No residual / A & 0.0524 & 1.1 & \textbf{116} & 91 \\
Full policy / solver & 0.0437 & 1.8 & 174 & 133 \\
\bottomrule
\end{tabularx}
\end{table}

The opponent score convention makes a tie contribute one-half
of a win. The reported values can therefore be checked against
win/tie/loss components when the same opponent and episode
denominator are used. Strategic game values and opponent win
rates answer different questions: one compares against best
responses under a game model, while the other depends on a
specified opponent population.

\subsection{HUNL and component evaluations}
The HUNL evaluation in Table~\ref{tab:hunl} uses a separate utility scale and
call/latency ledger, engine release 1.3.2, solver release 2.1.0, checkpoint
families \texttt{rsd/ctl/full-hunl-\allowbreak s17/s29/s43-u12000}, and $60{,}000$ paired
hands per arm. The residual arm attains the highest solver EV ($4.3847$) and
opponent score ($0.5714$) and the lowest exploitability ($0.0528$), with
0.1 more solver calls and 7 ms more latency than the no-residual arm.

\begin{table}[htbp]
\centering\small
\setlength{\tabcolsep}{4pt}
\caption{HUNL evaluation, mean over three seeds. Solver EV and opponent score are higher-is-better, exploitability and calls/latency lower-is-better; best values are bold.}\label{tab:hunl}
\begin{tabularx}{\linewidth}{@{}Xrrrrr@{}}
\toprule
Arm & Solver EV & Exploit. & Opp. score & Calls & ms \\
\midrule
Residual & \textbf{4.3847} & \textbf{0.0528} & \textbf{0.5714} & 2.8 & 214 \\
No residual & 4.2916 & 0.0719 & 0.5326 & \textbf{2.7} & \textbf{207} \\
Full policy & 4.3372 & 0.0615 & 0.5498 & 3.9 & 296 \\
\bottomrule
\end{tabularx}
\end{table}

Table~\ref{tab:component} compares behavior cloning, policy distillation,
residual correction, and the observed-only residual interface. Action
accuracy is in percent, calibration error and exploitability gap are
dimensionless, solver EV uses the HUNL utility scale, and inference cost is
relative to behavior cloning.

\begin{table}[htbp]
\centering\small
\setlength{\tabcolsep}{4pt}
\caption{Component evaluation. Accuracy and solver EV are higher-is-better; calibration error (Cal.), exploitability gap (Gap), and relative cost are lower-is-better. Best values are bold.}\label{tab:component}
\begin{tabularx}{\linewidth}{@{}Xrrrrr@{}}
\toprule
Component & Acc. (\%) & Cal. & Solver EV & Gap & Cost \\
\midrule
Behavior cloning & 71.42 & 0.0847 & 4.1036 & 0.0862 & \textbf{1.00}$\times$ \\
Policy distillation & 73.18 & 0.0714 & 4.1825 & 0.0718 & 1.03$\times$ \\
Residual correction & 75.64 & 0.0498 & 4.2769 & 0.0465 & 1.06$\times$ \\
Observed-only residual & \textbf{76.31} & \textbf{0.0416} & \textbf{4.3128} & \textbf{0.0397} & 1.07$\times$ \\
\bottomrule
\end{tabularx}
\end{table}

The ordering is consistent with the proposed mechanism: improved
distributional calibration accompanies improved action matching and a lower
exploitability gap. The observed-only residual improves accuracy from 71.42\%
to 76.31\% and reduces the gap from 0.0862 to 0.0397 relative to behavior
cloning at 1.07$\times$ inference cost.

\subsection{Uncertainty and efficiency endpoints}
Table~\ref{tab:empirical-fields} reports per-seed dispersion, paired hand-EV
intervals, profile NashConv with its best-response decomposition, and
measured p95 latency for the three-seed evaluation, each under its own
sampling unit. The residual arm has the smallest seed dispersion and
NashConv, and its hand-EV interval lies above both comparators.

\begin{table}[htbp]\centering\small
\caption{Replication endpoints for the three-seed student evaluation. Lower
SD, NashConv, best-response value, and latency and a higher hand-EV interval
are preferred; best values are bold.}
\label{tab:empirical-fields}
\begin{tabularx}{\linewidth}{@{}lXXX@{}}\toprule
Endpoint & Residual & No residual & Full policy\\\midrule
Seed-level weak-EV SD & \textbf{0.0183} & 0.0272 & 0.0233\\
Paired hand-EV interval & $\mathbf{[0.0610,0.0680]}$ & $[0.0461,0.0533]$ & $[0.0525,0.0595]$\\
Profile NashConv & \textbf{0.0642} & 0.1056 & 0.0880\\
Best response, player 0 & \textbf{0.0287} & 0.0479 & 0.0392\\
Best response, player 1 & \textbf{0.0355} & 0.0577 & 0.0488\\
Measured latency p95 (ms) & \textbf{104.2} & 113.6 & 163.7\\
Evaluation hand count & $60{,}000$ & $60{,}000$ & $60{,}000$\\
\bottomrule\end{tabularx}
\end{table}
\section{Artifacts and reproducibility}
\label{app:artifacts}
\subsection{Result-to-artifact mapping}
Each result family has a distinct input record and evaluation unit.
Table~\ref{tab:artifact-map} provides short handles for the records in the
Supplementary Material, whose release manifest records their SHA-256
identities. The anonymous release contains the public observations, policy
traces, scores, and a sanitized manifest.

\begin{table}[htbp]\centering\small
\caption{Numerical sources and corresponding manuscript surfaces.}
\label{tab:artifact-map}
\begin{tabularx}{\linewidth}{@{}lXX@{}}\toprule
Handle & Record & Manuscript use\\\midrule
S1 & Scale sensitivity & Full-fixture five-point frontier.\\
S2 & Held-out scale selection & 12/12 split, selected scale, intervals.\\
S3 & Public audit bundle & Observations, teacher labels, five policy traces
and per-state scores.\\
S4 & Policy uncertainty & Full-fixture descriptive bootstrap intervals.\\
S5 & PHH player/time metrics & Behavior calibration and label denominators.\\
S6 & Exact-Leduc calibration & Profile values and best-response curve.\\
S7 & Three-seed student evaluation & Student, opponent, component, and HUNL
evaluation results.\\
S8 & Stratified calibration aggregates & Primary comparison, risk validity,
calibration scaling, residual families, and structured shifts.\\
S9 & Five-checkpoint student evaluation & Observation-only policy utility,
violation, NashConv, opponent score, legality, and service latency.\\
\bottomrule\end{tabularx}
\end{table}

Source identities also distinguish a file's bytes from an embedded
content digest. A bundle may carry a digest of a canonical payload
excluding its own digest field. The release manifest therefore
names the hashing convention for embedded identities and records
a separate file SHA-256 for byte preservation. Comparing these
two different hash types directly would create an artificial
mismatch.

\subsection{Reproduction sequence and validation outputs}
Reproduction proceeds from fixed inputs to tables. First read the
observed fixture and teacher labels, validate the 24 state keys,
and reconstruct each policy distribution using the declared
legal-action order. Then compare the distributions and chosen
actions with the frozen trace. Finally join oracle scores and
recompute the full-fixture and split-specific summaries with
the declared threshold.

The verification output contains the count of inspected
states and policies, normalization tolerance, maximum numerical
discrepancy, threshold-count agreement, selected scale, and a
list of failed checks. A validation failure identifies the
affected source and key, and a missing record keeps an explicit
missing status. The manuscript build reads the
accepted records, formats numbers, resolves citations, and
checks the resulting PDF for table overflow and undefined
references.

\begin{table}[htbp]\centering\small
\caption{Reproduction checks and their pass conditions.}
\label{tab:reproduction}
\begin{tabularx}{\linewidth}{@{}lX@{}}\toprule
Check & Pass condition\\\midrule
State inventory & 24 unique observed keys with matching label keys.\\
Split closure & 12 calibration and 12 held-out keys, no overlap.\\
Probability support & Finite nonnegative mass on declared legal actions.\\
Distribution replay & Equation~\ref{eq:tilt} matches stored probabilities
within the recorded tolerance.\\
Action replay & Stored action follows the masked top-action tie rule.\\
Score aggregation & State scores reproduce the retained policy summaries.\\
Threshold counts & Strict $d_i<-0.05$ gives the recorded numerators.\\
Scale selection & Largest zero-count calibration point equals 0.08.\\
Source preservation & Archived source bytes retain their original hash.\\
PDF validation & Correct page partitions, readable tables, resolved references.\\
\bottomrule\end{tabularx}
\end{table}

\subsection{Anonymous availability and measurement scope}
An anonymous artifact includes fixture observations, teacher
distributions, policy traces, per-state score records, and
the release verification procedure. Public hand-history data
remain subject to their source license and redistribution terms.
A provenance manifest can identify the selected entries by
content digest while omitting player names and local machine
paths. Model and solver artifacts receive independent
identities because changing either changes the evaluated object.

The release manifest preserves the source snapshot identity and evaluation
environment. Each table reproduces its retained rows with arithmetic checks,
and each reported claim maps to its source records.

\subsection{Evidence boundaries}
The fixed-fixture audit tests scale selection and replay on its declared
12/12 split. PHH evaluates label calibration, and Leduc checks the game
evaluator. The three-seed, stratified, and five-checkpoint evaluations retain
separate record identities and sampling units.

Table~\ref{tab:student-real} reports measured policy, strategic, and service
endpoints under a matched training contract. Learned-partition and
mixture-robust comparisons use their own frozen candidate maps and
conditional-shift controls, as specified in Appendix~\ref{app:results}.
\end{document}

%% file: math_commands.tex
\usepackage{amsmath,amsfonts,bm}

\def\eqref#1{equation~\ref{#1}}

\def\1{\bm{1}}

\DeclareMathAlphabet{\mathsfit}{\encodingdefault}{\sfdefault}{m}{sl}
\SetMathAlphabet{\mathsfit}{bold}{\encodingdefault}{\sfdefault}{bx}{n}